\documentclass[10pt,twocolumn,letterpaper]{article}

\usepackage{cvpr}              

\usepackage{needspace}

\usepackage[ruled,vlined,linesnumbered]{algorithm2e}

\usepackage[noend]{algpseudocode} 

\usepackage{graphicx}
\graphicspath{{figures/}}

\usepackage{multirow}
\usepackage{tabularx}
\usepackage{booktabs}
\usepackage{adjustbox}

\definecolor{cvprblue}{rgb}{0.21,0.49,0.74}
\usepackage[pagebackref,breaklinks,colorlinks,allcolors=cvprblue]{hyperref}
\usepackage{longtable}
\usepackage{booktabs}
\usepackage{array}

\title{BootOOD: Self-Supervised Out-of-Distribution Detection via Synthetic Sample Exposure under Neural Collapse}

\author{
Yuanchao Wang\textsuperscript{1} \quad
Tian Qin\textsuperscript{1} \quad
Eduardo Valle\textsuperscript{2} \quad
Bruno Abrahao\textsuperscript{1,3} \\
\textsuperscript{1}NYU Shanghai Center for Data Science \\
\textsuperscript{2}Intercom \\
\textsuperscript{3}Leonard N.\ Stern School of Business, New York University \\
{\tt\small yw6570@nyu.edu \quad tq2067@nyu.edu \quad valle.do.eduardo@gmail.com \quad abrahao@nyu.edu}
}

\begin{document}
\maketitle

\begin{abstract}
Out-of-distribution (OOD) detection is critical for deploying image classifiers in safety-sensitive environments, yet existing detectors often struggle when OOD samples are semantically similar to the in-distribution (ID) classes. We present \textbf{BootOOD}, a fully self-supervised OOD detection framework that bootstraps exclusively from ID data and is explicitly designed to handle semantically challenging OOD samples. BootOOD synthesizes pseudo-OOD features through simple transformations of ID representations and leverages Neural Collapse (NC), where ID features cluster tightly around class means with consistent feature norms. Unlike prior approaches that aim to constrain OOD features into subspaces orthogonal to the collapsed ID means, BootOOD introduces a lightweight auxiliary head that performs radius-based classification on feature norms. This design decouples OOD detection from the primary classifier and imposes a relaxed requirement: OOD samples are learned to have smaller feature norms than ID features, which is easier to satisfy when ID and OOD are semantically close. Experiments on CIFAR-10, CIFAR-100, and ImageNet-200 show that BootOOD outperforms prior post-hoc methods, surpasses training-based methods without outlier exposure, and is competitive with state-of-the-art outlier-exposure approaches while maintaining or improving ID accuracy.

\end{abstract}

\section{Introduction}

Deep neural networks achieve impressive accuracy on large-scale image classification benchmarks, yet their predictions can be brittle when inputs deviate from the training distribution. In safety-sensitive applications such as autonomous driving, medical diagnosis, and open-world recognition, this vulnerability can lead to high-confidence errors that undermine reliability. Out-of-distribution (OOD) detection addresses this issue by equipping a classifier with the ability to detect inputs that differ from in-distribution (ID) training data and abstain or defer decisions when necessary~\cite{Hendrycks2017MSP,Guo2017Calibration,Hendrycks21Faces}. Despite intensive progress, reliably detecting OOD samples remains challenging, especially when OOD data are semantically similar to ID classes~\cite{Yang2021SCOOD,DSAAhmed}.

Early work on OOD detection focused on post-hoc scoring rules applied to a frozen classifier. The simplest baseline uses the maximum softmax probability (MSP) as an OOD score~\cite{Hendrycks2017MSP}, and subsequent methods refine this idea through calibration~\cite{Guo2017Calibration}, feature-space distances~\cite{Ren2018Mahalanobis}, energy-based scores~\cite{Liu20Energy}, gradient statistics~\cite{Huang2021GradNorm,abbas2025MedianofGradients}, activation rectification~\cite{Sun2021ReAct}, nearest neighbors~\cite{Sun2022DeepKNN}, sparsification~\cite{Sun2022DICE}, or representation manipulation~\cite{Wang2022ViM,Song2022RankFeat,Djurisic2023ASH,Zhang2023SHE}. These post-hoc methods are attractive because they can be deployed on off-the-shelf classifiers without modifying training. However, large-scale evaluations have shown that they often saturate under realistic and fine-grained distribution shifts~\cite{Hendrycks2022ScalingOOD,Yang22OpenOODv1,Zhang24OpenOODv15,Bitterwolf2023NINCO}, particularly when OOD samples share visual semantics with ID classes rather than being trivially far from the training distribution.

To overcome these limitations, recent work has explored training-based OOD detectors that modify the learning objective. Some approaches adjust the classification loss to enforce calibrated confidence or explicit reject options~\cite{Lee18ConfOOD,DeVries2018Confidence,Wei2022LogitNorm}, while others introduce auxiliary heads or embedding constraints~\cite{Ming2023CIDER,Tao2023NPOS,Ming22POEM,Hsu2020GODIN}. A dominant line of work uses outlier exposure (OE)~\cite{Hendrycks2019OE}, where external datasets such as synthetic noise, texture datasets, or large-scale image collections are treated as OOD and included during training. OE and its variants with virtual or synthesized outliers~\cite{Du2022VOS,Zhang2023MixOE,Wang23LAD,Sun23SROOD} can significantly improve robustness, but they inherit two important limitations. First, 
external OOD datasets often do not comprehensively reflect the OOD samples encountered at test time, creating an OOD-distribution mismatch that limits generalization. Second, encouraging the classifier to carve out decision regions that separate ID classes from a wide variety of outliers can be detrimental to in-distribution accuracy and may still fail when encountering near-OOD samples that closely resemble ID classes~\cite{Hendrycks2022ScalingOOD,Yang2021SCOOD,Bitterwolf2023NINCO}.

\begin{figure}[t]
    \centering
    \includegraphics[width=1\linewidth]{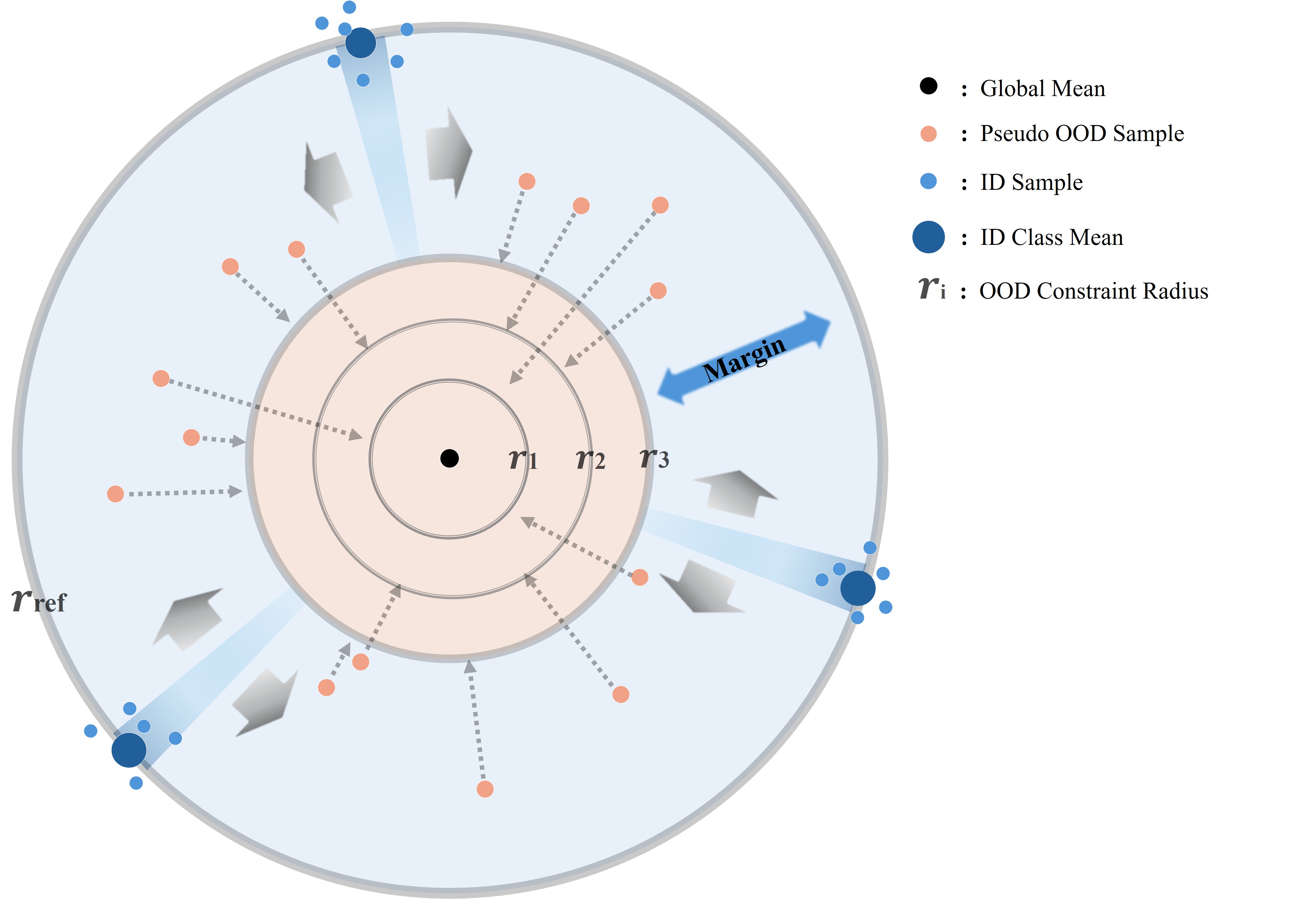}
    \caption{\textbf{Overview of the BootOOD geometry.} 
ID class features collapse around their class means with consistent norms under Neural Collapse. 
BootOOD synthesizes pseudo-OOD features in between these means and trains an auxiliary radius-based head to classify them as having smaller feature norms. 
The global mean defines the center, ID samples occupy a reference radius $r_{\text{ref}}$, and a sequence of inner radii $r_1, r_2, r_3$ specifies relaxed OOD constraint regions. 
By separating ID and OOD purely in feature-norm space, BootOOD avoids enforcing strict angular (orthogonality) constraints and relieves the main classifier from modeling OOD structure.}
    \label{fig:overview}
\end{figure}

Semantically similar OOD detection is especially challenging on benchmarks where ID classes occupy a dense semantic manifold. Examples include fine-grained or hierarchical label spaces such as CIFAR-100~\cite{Krizhevsky2009CIFAR}, ImageNet-200 and related ImageNet-based splits~\cite{Ridnik2021ImageNet21K,Bitterwolf2023NINCO}, or semantically coherent near-OOD benchmarks such as SCOOD~\cite{Yang2021SCOOD}. In these settings, OOD images often come from visually related categories, for example unseen bird species or vehicles, rather than from clearly different domains such as textures~\cite{Cimpoi2014DTD} or scenes~\cite{Zhou2018Places}. OpenOOD and its extensions~\cite{Yang22OpenOODv1,Zhang24OpenOODv15} highlight that many existing methods perform well on far-OOD settings yet struggle when ID and OOD classes share similar semantics. Designing detectors that explicitly target these semantically challenging regimes without sacrificing ID performance remains an open problem.

In this work, we revisit training-based OOD detection through the lens of Neural Collapse (NC)~\cite{Papyan2020NC}. NC describes an empirical phenomenon that emerges in the terminal phase of training large classifiers, where penultimate-layer features of each class cluster tightly around their class mean, class means form an approximately equiangular tight frame, and the classifier weights align with these means. An often underexploited aspect of NC is that the norms of ID class means and features become highly consistent across classes. Recent methods have tried to exploit NC structure by enforcing geometric constraints on OOD data during training, for example pushing them into subspaces that are orthogonal to the collapsed ID class means or enforcing large angular separation~\cite{Ammar2024NECO,Wu2025SepLoss,Liu24NC,Haas23LinkingL2NC,Park23FeatureNorm}. While effective in some regimes, these orthogonality-style constraints can be overly strict when ID and OOD classes are semantically similar, since realistic near-OOD samples may lie in between ID class directions rather than in strictly orthogonal subspaces.

We introduce \textbf{BootOOD}, a fully self-supervised OOD detection framework that is designed specifically for semantically similar OOD scenarios. The key idea is to relieve the main classifier from the burden of accommodating OOD data during training and to move OOD modeling into a lightweight auxiliary head that operates solely on feature norms. BootOOD bootstraps synthetic pseudo-OOD features by interpolating ID representations that lie between collapsed class means, then trains an auxiliary radius-based classifier to regard these pseudo-OOD features as having shorter norms than ID features. This norm-based separation is a relaxed requirement compared to enforcing geometric orthogonality between ID and OOD, and it leverages the consistent norms induced by NC to define a simple radial decision rule.

Figure~\ref{fig:overview} illustrates the BootOOD geometry. ID class means form an approximate equiangular configuration around a global mean in the feature space, and ID samples concentrate near a reference radius that reflects the collapsed feature norm. BootOOD defines a sequence of inner radii that specify constraint regions for pseudo-OOD features. Synthetic OOD samples are generated between class means and trained to occupy these smaller radii, creating a margin in feature norm between ID and OOD while leaving the angular structure of the classifier largely untouched. Because the auxiliary head is decoupled from the primary classifier, BootOOD preserves ID accuracy and adds negligible computational overhead at inference.

We evaluate BootOOD on CIFAR-10, CIFAR-100, and ImageNet-200 under standard and semantically challenging OOD benchmarks~\cite{Krizhevsky2009CIFAR,Yang2021SCOOD,Yang22OpenOODv1,Zhang24OpenOODv15,Bitterwolf2023NINCO}. BootOOD consistently improves OOD detection performance in near-OOD regimes while maintaining or improving ID accuracy. In particular, it outperforms prior post-hoc methods, surpasses training-based methods that do not rely on outlier exposure, and is competitive with state-of-the-art OE-based approaches, without requiring any real OOD data.

Our main contributions are as follows:
\begin{itemize}
    \item We propose \textbf{BootOOD}, a self-supervised, outlier-free OOD detection framework that decouples OOD modeling from the primary classifier through a lightweight auxiliary head operating on feature norms, offering a more stable alternative to prior post-hoc and training-based approaches in semantically similar OOD regimes.
    \item We leverage Neural Collapse to formulate a relaxed norm-based separation between ID and OOD features, and introduce a bootstrapping strategy that synthesizes pseudo-OOD features lying between collapsed class means, avoiding the strict geometric (e.g., orthogonality) constraints required by recent NC-based methods.
    \item We demonstrate that BootOOD delivers strong OOD performance on CIFAR-10, CIFAR-100, and ImageNet-200, outperforming prior post-hoc and outlier-free training-based methods and remaining competitive with outlier exposure approaches, while maintaining ID accuracy.
\end{itemize}

\section{Related Work}

\subsection{Post-hoc OOD Detection}

Post-hoc OOD detection methods operate on a fixed classifier and design scoring rules to distinguish ID and OOD samples at test time. The MSP baseline~\cite{Hendrycks2017MSP} uses the maximum softmax probability as an uncertainty score, and calibration techniques~\cite{Guo2017Calibration} further adjust confidence without changing the underlying classifier. Distance-based approaches compute scores from penultimate features, for example Mahalanobis distances to class-conditional Gaussians~\cite{Ren2018Mahalanobis} or Gram-matrix statistics~\cite{Sastry2020Gram}. Energy-based methods~\cite{Liu20Energy} interpret OOD detection in terms of the log-sum-exp of logits and can be combined with various training objectives.

Beyond logits and distances, other post-hoc approaches exploit gradients~\cite{Huang2021GradNorm}, activation rectification~\cite{Sun2021ReAct}, virtual-logit matching~\cite{Wang2022ViM}, deep nearest neighbors~\cite{Sun2022DeepKNN}, sparsification~\cite{Sun2022DICE}, rank-one feature removal~\cite{Song2022RankFeat}, or simple activation shaping~\cite{Djurisic2023ASH}. Hopfield-style energy models~\cite{Zhang2023SHE} and confidence branches~\cite{DeVries2018Confidence} provide additional alternatives. These methods typically require no extra training and are widely adopted due to their simplicity and compatibility with pretrained models.

However, comprehensive evaluations on large-scale and semantically rich benchmarks~\cite{Hendrycks2022ScalingOOD,Yang22OpenOODv1,Zhang24OpenOODv15,Bitterwolf2023NINCO} show that post-hoc methods often plateau when OOD samples are semantically similar to ID classes. In particular, near-OOD benchmarks such as SCOOD~\cite{Yang2021SCOOD} and ImageNet-based settings that use fine-grained or hierarchically related OOD classes~\cite{Ridnik2021ImageNet21K,Bitterwolf2023NINCO} reveal that many post-hoc scores are sensitive to the underlying representation and struggle to separate overlapping semantic manifolds.

\subsection{Training-based OOD Detection Without Outlier Exposure}

Training-based OOD detectors modify the learning objective in order to produce representations or logits that are intrinsically more amenable to OOD detection. Confidence-calibrated classifiers~\cite{Lee18ConfOOD,DeVries2018Confidence} encourage low confidence on hard or adversarial examples and provide explicit reject options. LogitNorm~\cite{Wei2022LogitNorm} normalizes logits to mitigate overconfidence and improves the calibration of decision boundaries. Methods based on hyperspherical embeddings~\cite{Ming2023CIDER} or nonparametric synthesis~\cite{Tao2023NPOS,Ming22POEM} design geometric or sampling constraints that separate ID and implicitly defined OOD directions without relying on external outlier datasets. Self-supervised learning has also been used to improve robustness and uncertainty estimation~\cite{Hendrycks2019SSL}. Unlike prior work that leverages self-supervision primarily to enhance in-distribution feature learning, our BootOOD framework uses self-supervision to explicitly model OOD data. Instead of enriching ID representations, we generate and refine OOD-aware supervisory signals that guide the network to better characterize, separate, and detect out-of-distribution samples.

These approaches avoid the practical difficulties of collecting and maintaining external OOD corpora, and they are often more stable when ID labels are fixed. Yet when semantically similar OOD samples are considered, the training objective may still place a heavy burden on the primary classifier. Enforcing complex geometry on top of the classification loss can distort decision regions and lead to an unfavorable trade-off between ID accuracy and OOD performance~\cite{Fang22LearnableOOD,Bitterwolf22BreakingDown}. BootOOD falls into this category of outlier-free training-based methods, but it explicitly decouples OOD modeling from the main classifier through an auxiliary norm-based head.

\subsection{Training-based Methods with Outlier Exposure}

Outlier exposure~\cite{Hendrycks2019OE} trains the classifier with an additional loss on external OOD data, encouraging uniform or low-confidence predictions on outlier samples. OE has inspired a variety of extensions that either refine the choice of outlier datasets or synthesize virtual outliers. VOS~\cite{Du2022VOS} learns virtual outlier synthesis in feature space. MixOE~\cite{Zhang2023MixOE} and related methods~\cite{Wang23LAD,Sun23SROOD} construct mixture or repaired OOD samples that better cover the semantic space. Other works integrate OE with large-scale robustness techniques such as AugMix~\cite{Hendrycks2020AugMix} or specialize to particular evaluation settings~\cite{Huang2021MOS,Hendrycks2022ScalingOOD}.

While OE-style methods can achieve strong performance, their reliance on external outlier collections raises questions about coverage and domain mismatch~\cite{Hendrycks21Faces,Hendrycks2022ScalingOOD}. In near-OOD regimes, even carefully chosen OE datasets may not align with the fine-grained semantics of the ID task, and the need to accommodate a wide variety of outliers within the primary classifier can degrade ID accuracy. In contrast, BootOOD does not require any real OOD data and instead bootstraps pseudo-OOD features directly from ID representations.

\subsection{Neural Collapse and Feature Norm for OOD Detection}

Neural Collapse (NC) has emerged as a unifying perspective on the geometry of deep classifiers in the terminal phase of training~\cite{Papyan2020NC,Vignesh2023NCreview}.
It characterizes a coupled set of phenomena where within-class features collapse to class means, centered class means approach a simplex equiangular tight frame (simplex ETF), last-layer weights align with these means up to scaling (self-duality), and the resulting classifier behaves similarly to a nearest-class-center rule in feature space. An important consequence of this geometry is the strong \emph{consistency of feature norms} across classes, yielding near-spherical and class-balanced ID clusters~\cite{Vignesh2023NCreview}.

Beyond explaining training dynamics, NC has motivated OOD and open-set recognition methods that exploit the structure of collapsed features and classifier weights.
NECO~\cite{Ammar2024NECO}, feature separation approaches~\cite{Wu2025SepLoss}, and NC-inspired detectors~\cite{Liu24NC,Haas23LinkingL2NC} enforce geometric constraints on synthetic or real OOD samples, often by encouraging them to lie in subspaces orthogonal to ID class means or by manipulating feature norms.
Park et al.~\cite{Park23FeatureNorm} further analyze the role of feature norms for OOD detection and show that norm statistics carry useful information about distributional shifts.

Complementary to NC-inspired geometry, between-class interpolation techniques such as Between-Class learning~\cite{tokozume2018between} and mixup~\cite{zhang2018mixup} synthesize samples that lie between ID classes.
Such interpolations are known to populate low-density, decision-boundary-adjacent regions in representation space and can serve as structured surrogates for ambiguous or hard examples.

These lines of work highlight that neural-collapse--induced structure can serve as a powerful inductive bias for OOD detection, yet many existing approaches still couple OOD constraints directly to the classifier and rely on strong angular-separation assumptions between ID and OOD samples~\cite{Ammar2024NECO,Wu2025SepLoss}.

\paragraph{Our differences.}
BootOOD differs in two key aspects.
First, it leverages the consistent norms of NC to formulate a relaxed, purely radial separation between ID and OOD, rather than enforcing strict orthogonality.
Second, it delegates OOD modeling to an auxiliary radius-based head that consumes both ID and bootstrapped pseudo-OOD features, which reduces interference with the classifier responsible for ID accuracy.
As we show in our experiments, this design is particularly effective for semantically similar OOD benchmarks on CIFAR-10, CIFAR-100, and ImageNet-200.

\section{Methodology}
\label{sec:method}

\subsection{Problem Setup}

We consider supervised image classification with inputs
\(x \in \mathbb{R}^{d}\) and categorical labels \(y \in \{1,\ldots,C\}\).
A backbone network \(f_{\theta} : \mathbb{R}^{d} \!\to\! \mathbb{R}^{m}\)
extracts penultimate-layer features
\[
h(x) = f_{\theta}(x),
\]
and a linear classifier
\(W = [w_1,\ldots,w_C]^\top \in \mathbb{R}^{C \times m}\)
produces logits
\[
z(x) = W h(x).
\]
Training uses only in-distribution (ID) data; no real OOD samples are used.

To simplify notation, we denote
\[
\mathrm{norm}(u) = \frac{u}{\|u\|_2},
\]
and we use $\hat{u} = \mathrm{norm}(u)$ for L2-normalized vectors.

\subsection{Neural Collapse as a Geometric Prior for BootOOD}

\paragraph{NC essentials and geometry.}
We build on the empirical observation that modern deep classifiers often enter a Neural Collapse (NC) regime in the terminal phase of training.
In this regime, penultimate features concentrate around class means, while centered class means form an approximate simplex ETF and last-layer weights align with them (self-duality)~\cite{Papyan2020NC,Vignesh2023NCreview}.
Concretely, letting $\mu_c \in \mathbb{R}^d$ denote the mean feature of class $c$ and $\tilde{\mu}_c$ its centered version,
\begin{equation}
\label{eq:nc_etf}
\begin{aligned}
\tilde{\mu}_c
&= \mu_c - \frac{1}{C}\sum_{j=1}^C \mu_j, \\
\|\tilde{\mu}_c\|
&\approx m, \\
\langle \tilde{\mu}_i,\tilde{\mu}_j\rangle
&\approx m^2\cos\theta,\quad i\neq j, \\
\cos\theta
&= -\frac{1}{C-1}, \\
w_c
&\approx \alpha\,\tilde{\mu}_c.
\end{aligned}
\end{equation}

where $C$ is the number of classes, $m>0$ and $\alpha>0$ are global scales, and $w_c$ is the $c$-th classifier weight.
A practically useful consequence is that ID feature norms become highly consistent across classes, producing near-spherical ID clusters~\cite{Vignesh2023NCreview}.

\paragraph{Phase-1 $\rightarrow$ Phase-2 warm start.}
BootOOD explicitly leverages this NC geometry.
In \textbf{Phase-1}, we train a standard classifier until it approaches the NC regime, and estimate a global ID center $\mu$ and a reference radius $r_{\text{ref}}$ using exponential moving averages:
\begin{equation}
\begin{aligned}
  \mu &\approx \mathbb{E}[h], \\
  r_{\text{ref}} &\approx \mathbb{E}\bigl[\|h-\mu\|\bigr],
\end{aligned}
\label{eq:center_radius}
\end{equation}
where $h$ denotes penultimate features of ID samples.
This yields a stable backbone geometry on which we attach an OOD-specific auxiliary head.

\paragraph{Why NC helps OOD detection.}
Under NC, ID data already forms tight, near-spherical clusters with consistent radii.
BootOOD converts this geometry into an inductive bias for OOD detection: pseudo-OOD features are driven towards \emph{smaller radii} (norm separation) and \emph{non-ID directions} (angular separation), while preserving the NC structure of the backbone and classifier.
This enlarged margin improves standard post-hoc scores (MSP, entropy, feature norm) and aligns with evidence that NC-induced geometry benefits OOD and open-set recognition~\cite{Papyan2020NC,Vignesh2023NCreview,Ammar2024NECO}.

\subsection{Radius head and directional separation.}

In \textbf{Phase-2}, we keep optimizing the backbone and the classifier using the standard
in-distribution objective, while attaching an auxiliary \emph{radius head}
$h_{\mathrm{rad}}:\mathbb{R}^d\to\mathbb{R}^K$
to model the radial structure of features relative to the global ID center~$\mu$.
Importantly, the classifier continues to be trained by the ID classification loss;
OOD-specific objectives are designed such that they do not directly alter the decision head.

Given a feature $h$, we consider its radius $\|h-\mu\|$ and define target radii
\begin{equation}
\begin{aligned}
  \{\rho_k\}_{k=1}^K &\subset [R_{\min},\,\rho_{\max}], \\
  \rho_{\max} &= (1-\gamma)\, r_{\mathrm{ref}},
\end{aligned}
\label{eq:rho_grid}
\end{equation}
linearly spaced between $R_{\min}$ and $\rho_{\max}<r_{\mathrm{ref}}$.
Here $\gamma\in(0,1)$ is a margin fraction that enforces pseudo-OOD features to lie strictly
inside the typical ID radius.

The radius head is trained using a combination of
(i) a $K$-way radial classification loss and
(ii) a regression term aligning $\|h-\mu\|$ to its designated target radius~$\rho_k$,
weighted by $\lambda_{\mathrm{cls}}$ and $\lambda_{\mathrm{mse}}$, respectively.

To complement radial separation, we additionally penalize the cosine similarity between pseudo-OOD features and classifier weights $\{w_c\}$.
Under the NC3 alignment in Eq.~(\ref{eq:nc_etf}), this discourages pseudo-OOD features from aligning with the ID simplex-ETF directions, providing an angular margin without modifying the backbone classifier.

Overall, the auxiliary radius head and the separation objective act as
\emph{training-time geometric regularizers} that reshape the representation
space beyond the ID manifold.
Importantly, they do not participate in inference.
At test time, the learned classifier is kept unchanged, and OOD detection
is performed using standard post-hoc scoring functions on the resulting
feature geometry.

\subsection{Pseudo-OOD Feature Generation}

Given an ID minibatch with penultimate features
\(\{h_i\}_{i=1}^B\) and labels \(\{y_i\}_{i=1}^B\),
we generate pseudo-OOD features directly in the feature space.
Two features \((h_i, h_j)\) from different ID classes are randomly paired, and we sample
\(\lambda \sim \mathrm{Beta}(\alpha,\alpha)\)
to form a mixed representation
\begin{equation}
h_{\mathrm{ood}}
= \lambda h_i + (1-\lambda) h_j,
\qquad y_i \neq y_j.
\label{eq:mixup}
\end{equation}
Thus, all pseudo-OOD samples used during training arise from
\emph{feature-level mixup of ID features}, without relying on any external
OOD data.

We denote the normalized pseudo-OOD feature as
\[
\tilde h = \hat{h}_{\mathrm{ood}}.
\]

\paragraph{Why Mixup-Generated Pseudo-OOD Targets Near-OOD.}
Near-OOD samples are known to lie in the \emph{low-density, between-class} region of the ID feature space and often behave as semantic interpolations between ID manifolds~\citep{zhang2018mixup,tokozume2018between}.
This region can be viewed as the complement of high-density class neighborhoods within the convex hull:
\[
\mathrm{Conv}\big(\{h_c\}\big)\setminus \bigcup_c \mathcal{N}(h_c).
\]
Feature-level mixup in Eq.~(\ref{eq:mixup}) guarantees that $h_{\mathrm{ood}} \in \mathrm{Conv}(\{h_c\})$, placing pseudo-OOD features exactly in this between-class zone.

\paragraph{Connections to Between-Class learning and standard mixup.}
Between-Class (BC) learning~\citep{tokozume2018between} mixes samples from different ID classes and regresses the mixing ratio, encouraging representations to populate structured division points between classes.
Standard mixup~\citep{zhang2018mixup} performs convex combinations (typically in input space) and uses soft labels for vicinal risk minimization.
In contrast, BootOOD applies a mixup-style operator \emph{directly in feature space} and \emph{never treats mixed features as ID}: we do not interpolate labels, and all mixed features are explicitly tagged as pseudo-OOD and supervised only by the auxiliary radius head.
This preserves standard ERM on genuine ID samples while injecting structured near-OOD supervision without external OOD data.

\paragraph{NC geometry and norm shrinkage under simplex ETF.}
Under Neural Collapse, centered class means form a simplex ETF with equal norm~\citep{Papyan2020NC,Vignesh2023NCreview}.
Let $\mu_c=\mathbb{E}[h\,|\,y=c]$ and define the global mean $\mu=\frac{1}{C}\sum_{j=1}^C \mu_j$, with centered means $\tilde{\mu}_c=\mu_c-\mu$.
In the NC regime, $\|\tilde{\mu}_c\|\approx m$ and $\langle\tilde{\mu}_i,\tilde{\mu}_j\rangle\approx m^2\cos\theta$ for $i\neq j$, with $\cos\theta=-\frac{1}{C-1}$.

Consider cross-class mixup at the (centered) class-mean level:
\[
\tilde{h}_{\mathrm{mix}}
=\lambda\,\tilde{\mu}_i + (1-\lambda)\,\tilde{\mu}_j,
\qquad i\neq j,\ \lambda\in(0,1).
\]
Then
\begin{equation}
\begin{aligned}
\|\tilde{h}_{\mathrm{mix}}\|^2
&= m^2\bigl[\lambda^2 + (1-\lambda)^2 + 2\lambda(1-\lambda)\cos\theta\bigr] \\
&= m^2\bigl[1 - 2\lambda(1-\lambda)(1-\cos\theta)\bigr].
\end{aligned}
\label{eq:mixup_norm_shrink}
\end{equation}
Because $1-\cos\theta>0$ for a simplex ETF, Eq.~(\ref{eq:mixup_norm_shrink}) implies $\|\tilde{h}_{\mathrm{mix}}\|<m$ whenever $\lambda\in(0,1)$.
Hence, between-class mixup features lie on simplex faces and exhibit strictly smaller centered norms than class centers.
Moreover, with $\lambda\sim\mathrm{Beta}(\alpha,\alpha)$, larger $\alpha$ concentrates $\lambda$ near $1/2$, pushing typical mixup features deeper into the between-class faces and further shrinking their centered norms.

\paragraph{From between-class geometry to radius supervision.}
Feature norm is a class-agnostic confidence proxy for OOD detection~\citep{Park23FeatureNorm}.
Combining this with the ETF analysis above, mixup-generated features are simultaneously (i) between-class in direction and (ii) smaller in radius relative to the global center $\mu$.
Our auxiliary radius head exploits exactly these cues: it assigns mixed features to \emph{inner} radius shells (smaller target radii) while keeping ID features near the reference radius estimated from the NC backbone.
Compared to BC learning, which regresses the mixing ratio, we only require pseudo-OOD to occupy low-norm, between-class shells---a relaxed constraint that is particularly suitable for semantically similar near-OOD settings.

\paragraph{Connection to feature-norm OOD scoring.}
Since feature norm correlates with confidence~\citep{Park23FeatureNorm}, mixup-induced norm shrinkage naturally strengthens norm-based OOD scoring and complements MSP/Entropy.
Empirically, this is consistent with NC-inspired post-hoc detectors such as NECO~\citep{Ammar2024NECO}.

\subsection{Tracking the ID Feature Geometry}

BootOOD leverages the Neural Collapse (NC) geometry that naturally emerges
during late-phase cross-entropy training.
To model this geometry, we maintain two exponential moving averages (EMAs):

\paragraph{ID feature mean}
\begin{equation}
\mu \leftarrow \beta_\mu \, \mu
  + (1-\beta_\mu)\,\overline{h}_{\mathrm{ID}} ,
\label{eq:mu}
\end{equation}

\paragraph{ID feature radius}
\begin{equation}
r_{\mathrm{ref}}
\leftarrow
\beta_r \, r_{\mathrm{ref}}
+ (1-\beta_r) \, \|h - \mu\|_2 .
\label{eq:rref}
\end{equation}

The scalar \(r_{\mathrm{ref}}\) captures the collapsed ID radius,
while \(\mu\) tracks the global ID center; $\beta_*$ are hyperparameters for the EMA.

\subsection{Radius-Based Organization of Pseudo-OOD Samples}

We discretize the interval \([0, r_{\mathrm{ref}})\) into \(K\) inner-shell
target radii
\[
0 < \rho_1 < \rho_2 < \cdots < \rho_K < r_{\mathrm{ref}},
\]
e.g., via uniform or cosine spacing.
A lightweight radius head
\(g_{\phi} : \mathbb{R}^{m} \to \mathbb{R}^{K}\)
maps \(\tilde h\) to a distribution over shells.

Given a sampled shell index \(s\), we optimize:

\paragraph{Radius classification}
\begin{equation}
\mathcal{L}_{\mathrm{cls}}
= \mathrm{CE}\!\left(g_{\phi}(\tilde h),\, s\right).
\label{eq:cls}
\end{equation}

\paragraph{Radius regression}
\begin{equation}
\mathcal{L}_{\mathrm{reg}}
= \left( \|\tilde h - \mu\|_2 - \rho_s \right)^2 .
\label{eq:reg}
\end{equation}

This encourages pseudo-OOD features to populate well-defined inner shells of
decreasing radius, separating them from the ID outer shell near
\(r_{\mathrm{ref}}\).

\subsection{Directional Separation}

To further decouple pseudo-OOD features from ID classifier directions,
we impose a soft angular-separation penalty:
\begin{equation}
\mathcal{L}_{\mathrm{sep}}
= \mathbb{E}_{c}\!
  \left[
    \, \left| \langle \hat{\tilde h}, \hat w_c \rangle \right| \,
  \right],
\label{eq:sep}
\end{equation}
computed with the classifier weights \(W\) detached.
This is motivated by recent findings that enforcing angular disentanglement
improves OOD detection~\cite{Wu2025SepLoss}.

\subsection{Total Objective}

ID samples are optimized with standard cross-entropy:
\[
\mathcal{L}_{\mathrm{CE}}
= \mathrm{CE}(z(x),\, y).
\]

We define the pseudo-OOD loss as
\[
\mathcal{L}_{\mathrm{OOD}}
= \lambda_{\mathrm{cls}} \mathcal{L}_{\mathrm{cls}}
+ \lambda_{\mathrm{reg}} \mathcal{L}_{\mathrm{reg}}.
\]

The full training objective is
\begin{equation}
\mathcal{L}
= \mathcal{L}_{\mathrm{CE}}
+ \lambda_{\mathrm{ood}}(t)\,\mathcal{L}_{\mathrm{OOD}}
+ \lambda_{\mathrm{sep}}(t)\,\mathcal{L}_{\mathrm{sep}},
\label{eq:total}
\end{equation}
where \(\lambda_{\mathrm{ood}}(t)\) and \(\lambda_{\mathrm{sep}}(t)\) follow simple
linear warm-up schedules.
The radius head parameters $\phi$ are introduced as a separate optimizer group.

\subsection{Training Algorithm}

Algorithm~\ref{alg:bootoo} summarizes the full BootOOD training procedure.
Each iteration updates the ID classifier using standard cross-entropy, while
simultaneously organizing mixed pseudo-OOD features into inner-radius shells
and enforcing angular separation.
\SetKwInput{KwInit}{Init}

\begin{algorithm}[t]
\caption{BootOOD Training Procedure}
\label{alg:bootoo}

\KwIn{Training set $\mathcal{D}_{\mathrm{ID}}$, backbone $f_\theta$, classifier $W$, radius head $g_\phi$}
\KwInit{EMA mean $\mu$, EMA radius $r_{\mathrm{ref}}$, schedules $\lambda_{\mathrm{ood}}(t)$ and $\lambda_{\mathrm{sep}}(t)$}

\For{each training iteration}{
    \tcc{ID forward pass}

    Sample minibatch $(x_i, y_i)_{i=1}^B \sim \mathcal{D}_{\mathrm{ID}}$\;
    Compute ID features $h_i = f_\theta(x_i)$ and logits $z_i = W h_i$\;
    Update EMA center $\mu$ using Eq.~\eqref{eq:mu}\;
    Update EMA radius $r_{\mathrm{ref}}$ using Eq.~\eqref{eq:rref}\;
    Compute ID loss $\mathcal{L}_{\mathrm{CE}} = \mathrm{CE}(z_i, y_i)$\;

    \tcc{Pseudo-OOD feature generation}
    \For{$k = 1$ \KwTo $M$}{
        Randomly pick $i \neq j$ from $\{1,\ldots,B\}$\;
        Sample $\lambda \sim \mathrm{Beta}(\alpha,\alpha)$\;
        $h_{\mathrm{ood}}^{(k)} = \lambda h_i + (1-\lambda) h_j$\;
        $\tilde h^{(k)} = \mathrm{norm}(h_{\mathrm{ood}}^{(k)})$\;
    }

    \tcc{Radius-based OOD modeling}
    Sample shell index $s \sim \{1,\ldots,K\}$\;
    Compute radius-classification loss $\mathcal{L}_{\mathrm{cls}}$ via Eq.~\eqref{eq:cls}\;
    Compute radius-regression loss $\mathcal{L}_{\mathrm{reg}}$ via Eq.~\eqref{eq:reg}\;

    \tcc{Directional separation}
    Compute angular-separation loss $\mathcal{L}_{\mathrm{sep}}$ via Eq.~\eqref{eq:sep}\;

    \tcc{Total loss and update}
    $\mathcal{L}
     = \mathcal{L}_{\mathrm{CE}}
     + \lambda_{\mathrm{ood}}(t)\big(
        \lambda_{\mathrm{cls}}\mathcal{L}_{\mathrm{cls}}
      + \lambda_{\mathrm{reg}}\mathcal{L}_{\mathrm{reg}}
      \big)
     + \lambda_{\mathrm{sep}}(t)\mathcal{L}_{\mathrm{sep}}$\;

    Update $\theta$, $\phi$, and $W$ using $\nabla \mathcal{L}$\;
}
\end{algorithm}

\begin{table}[t]
  \caption{Compatibility of Our Method w/ Various PostProcessors}
  \label{tab:abla_3in1_5score}
  \centering

  \setlength{\tabcolsep}{8pt}
  \renewcommand{\arraystretch}{1.12}

  \begingroup
  \fontsize{12pt}{15pt}\selectfont
  \resizebox{\linewidth}{!}{%
  \begin{tabular}{@{}l|cc|cc|cc@{}}
    \toprule
      \multirow{2}{*}{\textbf{Method}} &
      \multicolumn{2}{c|}{\textbf{CIFAR-10}} &
      \multicolumn{2}{c|}{\textbf{CIFAR-100}} &
      \multicolumn{2}{c}{\textbf{ImageNet-200}} \\
      & \textbf{FPR@95} & \textbf{AUROC}
      & \textbf{FPR@95} & \textbf{AUROC}
      & \textbf{FPR@95} & \textbf{AUROC} \\
    \midrule
      EBO\cite{Liu20Energy}
      & 64.19($\pm$0.67) & 87.65($\pm$0.53)
      & 53.17($\pm$0.57) & 81.98($\pm$0.53)
      & 63.16($\pm$0.45) & 80.79($\pm$0.38) \\
      Entropy\cite{Liu2023GEN}
      & 62.71($\pm$0.33) & 88.32($\pm$0.43)
      & \textbf{51.03}($\pm$0.36) & \textbf{83.09}($\pm$0.23)
      & 58.65($\pm$0.34) & 82.95($\pm$0.32) \\
      ReAct\cite{Sun2021ReAct}
      & 76.73($\pm$0.74) & 87.01($\pm$0.81)
      & 54.65($\pm$0.55) & 81.95($\pm$0.43)
      & 65.50($\pm$1.58) & 81.18($\pm$1.50) \\
      Norm\cite{Park23FeatureNorm, Yu2023FeatureNormBlock}
      & \textbf{31.34}($\pm$0.27) & \textbf{92.40}($\pm$0.28)
      & 98.83($\pm$1.08) & 43.09($\pm$1.23)
      & 79.11($\pm$0.83) & 74.75($\pm$0.80) \\
      MSP\cite{Hendrycks2017MSP}
      & 62.32($\pm$0.41) & 89.17($\pm$0.43)
      & 53.28($\pm$0.77) & 82.31($\pm$0.51)
      & \textbf{53.16}($\pm$0.22) & \textbf{84.16}($\pm$0.20) \\
    \bottomrule
  \end{tabular}
  }
  \endgroup
\end{table}

\subsection{Inference and OOD Scoring}
At test time we discard the radius head and use only the backbone and classifier. Given a test image, we obtain its logits and penultimate feature and compute five post-hoc OOD scores. We compare these five scores across datasets and OOD sets in Table.~\ref{tab:abla_3in1_5score}. Thresholds for binary ID/OOD decisions are chosen on a held-out ID validation split (e.g., a fixed percentile on ID scores) or following the OpenOOD evaluation protocol.

\section{Experiments}
\label{sec:expriments}

\subsection{Datasets, Protocols, and Metrics}

\paragraph{Protocol.}
We follow the standard \textbf{OpenOOD v1.5} protocol \citep{Zhang24OpenOODv15}: models are trained on \emph{ID-only} data and evaluated on held-out ID/OOD sets. All experiments use ResNet-18 with input resolution $32{\times}32$ for CIFAR-10/100 and $224{\times}224$ for ImageNet-200. Results are reported as mean~$\pm$~std over \textbf{3 seeds}, and initialization uses the corresponding OpenOOD pretrained checkpoints.

\paragraph{OOD sets and near/far rationale.}
For CIFAR-10 as ID, the near-OOD pool contains CIFAR-100 and Tiny-ImageNet (TIN) with overlapping classes removed \citep{Krizhevsky2009CIFAR,Yang2021SCOOD}; the far-OOD pool contains MNIST \citep{deng2012mnist}, SVHN \citep{Netzer2011SVHN}, Textures \citep{Cimpoi2014DTD}, and Places365 \citep{Zhou2018Places}.  
For CIFAR-100 as ID, the near-OOD pool is CIFAR-10 and TIN, and the far-OOD pool matches the CIFAR-10 setting.  
For ImageNet-200 as ID, following OpenOOD, the near-OOD pool uses SSB-Hard \cite{Vaze2022OSR} and NINCO \citep{Bitterwolf2023NINCO}; the far-OOD pool consists of iNaturalist \citep{VanHorn2018iNat}, Textures, and OpenImage-O \citep{Wang2022ViM}.

\paragraph{Data usage.}
\textbf{BootOOD uses ID-only training and ID-only validation.}  
Real OOD samples are never used during training; pseudo-OOD arises solely from our feature-level generator.  
When a baseline requires OOD validation (e.g., temperature scaling), we follow OpenOOD and draw a small disjoint OOD validation split from the designated OOD pool.

\paragraph{Evaluation.}
We validate using multiple post-processors—entropy\cite{Liu2023GEN}, norm\cite{Park23FeatureNorm,Yu2023FeatureNormBlock}, ReAct\cite{Sun2021ReAct}, EBO\cite{Liu20Energy}, and MSP\cite{Hendrycks2017MSP}.  
Because different datasets favor different scoring rules, we select the post-processor that achieves the best performance on the \emph{ID-only validation split}, and use that fixed scorer for final testing, as shown in Table \ref{tab:abla_3in1_5score}. The backbone is unchanged at test time; only the scoring rule differs. This protocol avoids tuning to a single scoring rule and provides a fair comparison across datasets.

\paragraph{Metrics.}
We report AUROC, FPR95, AUPR\(_\text{IN}\), AUPR\(_\text{OUT}\), and ID-ACC.  
Thresholds for FPR95/TNR95 follow OpenOOD and are determined on a held-out validation split. OOD test labels are never used for model selection.

\paragraph{Neural-Collapse warm-up.}
Our Phase-I training is designed to enter the \emph{terminal phase of training} (TPT),
where Neural Collapse (NC) is known to reliably emerge~\cite{Papyan2020NC,Vignesh2023NCreview}.
To this end, we continue ID-only training beyond zero classification error
before activating Phase-II regularization, ensuring that the feature geometry
has reached a stable NC regime.
Empirically, NC typically appears around $\sim70$ epochs on CIFAR-10,
$\sim80$ epochs on CIFAR-100, and $\sim100$ epochs on ImageNet-200.

\subsection{Baselines}
We follow the OpenOOD v1.5 taxonomy and compare it with representative methods with respect to performance and recency of publication from three categories: post-hoc inference, outlier-free training, and outlier-exposed training. The complete list of methods is presented in the Appendix.

\subsection{Main Results}

We compare \textbf{BootOOD} against representative methods from the three standard OOD categories in OpenOOD: \emph{post-hoc} methods, \emph{training-time} methods without real OOD, and \emph{training-time} methods with real OOD (four baselines per category). Following the OpenOOD protocol and its leaderboard focus, we report results on the \emph{near-OOD} setting only, which is widely regarded as the most challenging regime \cite{Yang22OpenOODv1,Zhang24OpenOODv15,Yang2021SCOOD,DSAAhmed}. For CIFAR-10/100 and ImageNet-200, all numbers in Table~\ref{tab:sota_3in1} are averaged over \textbf{3} random seeds.
\begin{table}[t]
  \captionsetup{skip=4pt}
  \caption{Near-OOD: CIFAR-10, CIFAR-100 and ImageNet-200.}
  \vspace{-2pt}
  \label{tab:sota_3in1}
  \centering

  \setlength{\tabcolsep}{1.6pt}
  \renewcommand{\arraystretch}{1.08}

  \resizebox{\linewidth}{!}{%
    \begin{tabular}{@{}l l c c c c c@{}}
      \toprule
      \textbf{Category} & \textbf{Method} &
      \shortstack[c]{\textbf{AUROC}} &
      \shortstack[c]{\textbf{FPR}\\\textbf{@95\%}} &
      \shortstack[c]{\textbf{AUPR}\\\textbf{IN}} &
      \shortstack[c]{\textbf{AUPR}\\\textbf{OUT}} &
      \textbf{ID-ACC} \\
      \midrule

      \multicolumn{7}{l}{\textbf{CIFAR-10} (ID Acc: 95.22\(\pm\)0.30)} \\
      \midrule
      Posthoc & SHE\cite{Zhang2023SHE}         & 80.84(\(\pm\)1.30) & 84.48(\(\pm\)1.10) & 75.53(\(\pm\)2.30) & 81.93(\(\pm\)1.39) & \textbf{95.22}(\(\pm\)0.30) \\
              & RMDS\cite{Ren2018Mahalanobis}  & 89.53(\(\pm\)0.35) & 42.19(\(\pm\)0.31) & 89.79(\(\pm\)0.26) & 87.48(\(\pm\)0.34) & \textbf{95.22}(\(\pm\)0.30) \\
              & KNN\cite{Sun2022DeepKNN}       & 90.70(\(\pm\)0.10) & 34.54(\(\pm\)0.22) & 91.73(\(\pm\)0.21) & 88.71(\(\pm\)0.36) & \textbf{95.22}(\(\pm\)0.30) \\
              & NECO\cite{Ammar2024NECO}       & 91.52(\(\pm\)2.63) & 37.32(\(\pm\)0.61) & 91.86(\(\pm\)1.42) & \textbf{90.23}(\(\pm\)1.01) & \textbf{95.22}(\(\pm\)0.30) \\
              & \textbf{BootOOD (Ours)}        & \textbf{92.40}(\(\pm\)0.28) & \textbf{31.34}(\(\pm\)0.27) & \textbf{93.75}(\(\pm\)0.36) & 89.94(\(\pm\)0.35) & 95.08(\(\pm\)0.43) \\
      \midrule
      Train (w/o) & NPOS\cite{Tao2023NPOS}                 & 83.31(\(\pm\)0.31) & 45.02(\(\pm\)0.42) & 86.82(\(\pm\)0.36) & 76.49(\(\pm\)0.27) & N/A \\
                  & PFS\cite{Wu2025SepLoss}                & 89.10(\(\pm\)1.25) & 37.38(\(\pm\)0.83) & 89.67(\(\pm\)0.96) & 81.01(\(\pm\)1.04) & 86.06(\(\pm\)0.54) \\
                  & ConfBranch\cite{DeVries2018Confidence} & 89.84(\(\pm\)0.24) & 31.97(\(\pm\)0.13) & 91.72(\(\pm\)0.29) & 85.84(\(\pm\)0.54) & 94.24(\(\pm\)0.11) \\
                  & CIDER\cite{Ming2023CIDER}              & 90.26(\(\pm\)0.20) & 32.94(\(\pm\)0.34) & 91.88(\(\pm\)0.18) & 87.72(\(\pm\)0.27) & N/A \\
                  & \textbf{BootOOD (Ours)}                & \textbf{92.40}(\(\pm\)0.28) & \textbf{31.34}(\(\pm\)0.27) & \textbf{93.75}(\(\pm\)0.36) & \textbf{89.94}(\(\pm\)0.35) & \textbf{95.08}(\(\pm\)0.43) \\
      \midrule
      Train (with) & MixOE\cite{Zhang2023MixOE}            & 88.57(\(\pm\)0.88) & 56.07(\(\pm\)0.74) & 86.43(\(\pm\)0.86) & 87.61(\(\pm\)0.69) & 94.56(\(\pm\)0.34) \\
                   & MCD\cite{Yu2019MCD}                   & 89.21(\(\pm\)0.14) & 25.65(\(\pm\)0.21) & 88.80(\(\pm\)0.18) & 91.21(\(\pm\)1.15) & 93.67(\(\pm\)0.04) \\
                   & UDG\cite{Yang2021SCOOD}               & 89.77(\(\pm\)0.24) & 36.18(\(\pm\)0.26) & 90.65(\(\pm\)0.19) & 87.05(\(\pm\)0.17) & 93.77(\(\pm\)0.87) \\
                   & OE\cite{Hendrycks2019OE}              & \textbf{94.14}(\(\pm\)0.21) & \textbf{19.65}(\(\pm\)0.33) & \textbf{94.57}(\(\pm\)0.28) & \textbf{93.88}(\(\pm\)0.19) & 94.24(\(\pm\)0.24) \\
                   & \textbf{BootOOD (Ours)}               & 92.40(\(\pm\)0.28) & 31.34(\(\pm\)0.27) & 93.75(\(\pm\)0.36) & 89.94(\(\pm\)0.35) & \textbf{95.08}(\(\pm\)0.43) \\

      \midrule

      \multicolumn{7}{l}{\textbf{CIFAR-100} (ID Acc: 77.19\(\pm\)0.13)} \\
      \midrule
      Posthoc & NECO\cite{Ammar2024NECO} & 47.91(\(\pm\)0.50) & 95.57(\(\pm\)0.20) & 50.94(\(\pm\)0.35) & 45.70(\(\pm\)0.32) & 71.35(\(\pm\)0.13) \\
              & SHE\cite{Zhang2023SHE}  & 78.24(\(\pm\)0.28) & 59.30(\(\pm\)0.18) & 82.15(\(\pm\)0.26) & 72.04(\(\pm\)0.30) & 77.09(\(\pm\)0.13) \\
              & TempScale\cite{Guo2017Calibration} & 81.46(\(\pm\)0.27) & 54.78(\(\pm\)0.07) & 84.03(\(\pm\)0.20) & 74.87(\(\pm\)0.22) & 77.43(\(\pm\)0.13) \\
              & MLS\cite{Hendrycks2022ScalingOOD}  & 81.56(\(\pm\)0.23) & 55.48(\(\pm\)0.07) & 83.88(\(\pm\)0.18) & 74.93(\(\pm\)0.22) & 77.19(\(\pm\)0.13) \\
              & \textbf{BootOOD (Ours)} & \textbf{83.09}(\(\pm\)0.23) & \textbf{51.03}(\(\pm\)0.36) & \textbf{84.93}(\(\pm\)0.24) & \textbf{77.65}(\(\pm\)0.28) & \textbf{78.89}(\(\pm\)0.12) \\
      \midrule
      Train (w/o) & PFS\cite{Wu2025SepLoss} & 72.19(\(\pm\)0.40) & 61.59(\(\pm\)0.30) & 77.04(\(\pm\)0.35) & 62.12(\(\pm\)0.40) & 54.39(\(\pm\)0.42) \\
                  & NPOS\cite{Tao2023NPOS} & 74.41(\(\pm\)0.37) & 67.11(\(\pm\)0.26) & 78.21(\(\pm\)0.24) & 66.72(\(\pm\)0.28) & N/A \\
                  & MOS\cite{Huang2021MOS} & 80.74(\(\pm\)0.18) & 55.06(\(\pm\)0.20) & 83.39(\(\pm\)0.18) & 73.87(\(\pm\)0.20) & 76.62(\(\pm\)0.20) \\
                  & VOS\cite{Du2022VOS}    & 81.02(\(\pm\)0.29) & 54.99(\(\pm\)0.18) & 83.81(\(\pm\)0.22) & 74.19(\(\pm\)0.24) & 77.09(\(\pm\)0.10) \\
                  & \textbf{BootOOD (Ours)} & \textbf{83.09}(\(\pm\)0.23) & \textbf{51.03}(\(\pm\)0.36) & \textbf{84.93}(\(\pm\)0.24) & \textbf{77.65}(\(\pm\)0.28) & \textbf{78.89}(\(\pm\)0.12) \\
      \midrule
      Train (with) & UDG\cite{Yang2021SCOOD}  & 77.63(\(\pm\)0.20) & 60.39(\(\pm\)0.22) & 81.06(\(\pm\)0.18) & 71.22(\(\pm\)0.24) & 71.93(\(\pm\)0.64) \\
                   & MCD\cite{Yu2019MCD}     & 79.72(\(\pm\)0.25) & 57.81(\(\pm\)0.30) & 82.05(\(\pm\)0.22) & 73.11(\(\pm\)0.24) & 74.98(\(\pm\)0.38) \\
                   & MixOE\cite{Zhang2023MixOE} & 79.88(\(\pm\)1.77) & 60.38(\(\pm\)0.95) & 67.40(\(\pm\)0.55) & \textbf{85.27}(\(\pm\)0.70) & 71.93(\(\pm\)0.64) \\
                   & OE\cite{Hendrycks2019OE} & 80.88(\(\pm\)0.20) & 55.04(\(\pm\)0.18) & 83.84(\(\pm\)0.16) & 74.89(\(\pm\)0.20) & 75.30(\(\pm\)0.42) \\
                   & \textbf{BootOOD (Ours)} & \textbf{83.09}(\(\pm\)0.23) & \textbf{51.03}(\(\pm\)0.36) & \textbf{84.93}(\(\pm\)0.24) & 77.65(\(\pm\)0.28) & \textbf{78.89}(\(\pm\)0.12) \\

      \midrule

      \multicolumn{7}{l}{\textbf{ImageNet-200} (ID Acc: 86.37\(\pm\)0.09)} \\
      \midrule
      Posthoc & NECO\cite{Ammar2024NECO}      & 55.89(\(\pm\)0.92) & 86.98(\(\pm\)0.58) & 45.81(\(\pm\)0.35) & 63.92(\(\pm\)0.42) & 86.37(\(\pm\)0.09) \\
              & SHE\cite{Zhang2023SHE}        & 80.18(\(\pm\)0.53) & 66.80(\(\pm\)0.32) & 64.25(\(\pm\)0.26) & 84.20(\(\pm\)0.24) & 86.37(\(\pm\)0.09) \\
              & TempScale\cite{Guo2017Calibration} & 83.24(\(\pm\)0.33) & 57.29(\(\pm\)0.32) & 69.04(\(\pm\)0.27) & \textbf{88.38}(\(\pm\)0.25) & 86.37(\(\pm\)0.09) \\
              & MSP\cite{Hendrycks2017MSP}    & 83.34(\(\pm\)0.38) & 54.82(\(\pm\)0.30) & \textbf{68.96}(\(\pm\)0.24) & 85.95(\(\pm\)0.25) & 86.37(\(\pm\)0.09) \\
              & \textbf{BootOOD (Ours)}       & \textbf{84.16}(\(\pm\)0.28) & \textbf{53.01}(\(\pm\)0.28) & 68.17(\(\pm\)0.30) & 86.78(\(\pm\)0.65) & \textbf{86.60}(\(\pm\)0.15) \\
      \midrule
      Train (w/o) & NPOS\cite{Tao2023NPOS}    & 74.41(\(\pm\)0.37) & 67.11(\(\pm\)0.26) & \textbf{78.21}(\(\pm\)0.24) & 66.72(\(\pm\)0.28) & N/A \\
                  & CIDER\cite{Ming2023CIDER} & 80.72(\(\pm\)1.38) & 61.40(\(\pm\)0.48) & 66.20(\(\pm\)0.38) & 80.70(\(\pm\)0.42) & N/A \\
                  & ARPL\cite{Chen2022ARPL}   & 82.10(\(\pm\)0.18) & 63.80(\(\pm\)0.34) & 67.70(\(\pm\)0.26) & 84.05(\(\pm\)0.28) & 84.10(\(\pm\)0.34) \\
                  & LogitNorm\cite{Wei2022LogitNorm} & 82.42(\(\pm\)0.22) & 57.06(\(\pm\)0.34) & 67.73(\(\pm\)0.26) & 86.22(\(\pm\)0.30) & 86.48(\(\pm\)0.18) \\
                  & \textbf{BootOOD (Ours)}   & \textbf{84.16}(\(\pm\)0.28) & \textbf{53.01}(\(\pm\)0.28) & 68.17(\(\pm\)0.30) & \textbf{86.78}(\(\pm\)0.65) & \textbf{86.60}(\(\pm\)0.15) \\
      \midrule
      Train (with) & UDG\cite{Yang2021SCOOD}  & 71.63(\(\pm\)1.52) & 66.42(\(\pm\)0.95) & 76.19(\(\pm\)0.72) & 66.73(\(\pm\)0.80) & 67.94(\(\pm\)1.39) \\
                   & MixOE\cite{Zhang2023MixOE} & 80.05(\(\pm\)0.07) & 59.61(\(\pm\)0.21) & \textbf{81.68}(\(\pm\)0.18) & 72.93(\(\pm\)0.20) & 85.01(\(\pm\)0.10) \\
                   & MCD\cite{Yu2019MCD}     & 81.42(\(\pm\)0.16) & 61.03(\(\pm\)0.27) & 79.85(\(\pm\)0.22) & 70.58(\(\pm\)0.25) & 85.12(\(\pm\)0.26) \\
                   & OE\cite{Hendrycks2019OE}& \textbf{85.02}(\(\pm\)0.24) & 53.16(\(\pm\)0.22) & 70.51(\(\pm\)0.22) & \textbf{86.85}(\(\pm\)0.26) & 85.90(\(\pm\)0.18) \\
                   & \textbf{BootOOD (Ours)} & 84.16(\(\pm\)0.28) & \textbf{53.01}(\(\pm\)0.28) & 68.17(\(\pm\)0.30) & 86.78(\(\pm\)0.65) & \textbf{86.60}(\(\pm\)0.15) \\

      \bottomrule
    \end{tabular}
  }

  \vspace{-8pt}
  \begin{flushleft}
    {\scriptsize
    CIDER\cite{Ming2023CIDER} and NPOS\cite{Tao2023NPOS} train the CNN backbone without the final linear classifier, and the official implementations do not provide code for evaluating ID accuracy.
    }
  \end{flushleft}
\end{table}

\paragraph{CIFAR-10.} \emph{BootOOD} achieves higher AUROC and lower FPR95 than all post-hoc methods and all training-time methods without outlier data. The largest gains appear on the \emph{near-OOD} split (CIFAR-100/TIN), where closeness to ID makes separation difficult.

\paragraph{CIFAR-100.} On this more fine-grained ID dataset, \emph{BootOOD} surpasses \emph{all} OpenOOD v1.5 post-hoc methods and \emph{all} training-time methods, \emph{including} those that leverage auxiliary outlier data (OE/MCD/UDG/MixOE). The advantage is especially pronounced on near-OOD (CIFAR-10/TIN), indicating that our radius-regularized representation improves class-conditional compactness and inter-class separation. Compared with the additional NECO and Feature Separation baselines, \emph{BootOOD} also leads consistently while preserving ID-ACC.

\paragraph{ImageNet-200.} Using the same ResNet-18 backbone with $224^2$ inputs, \emph{BootOOD} outperforms every OpenOOD v1.5 post-hoc method and all training-time methods \emph{without} outlier data. Improvements are most visible on near-OOD (SSB-Hard/NINCO) where many categories are fine-grained and visually similar to ID. We also find that \emph{BootOOD} is competitive with methods that rely on additional outliers even though it does not use them.

\paragraph{Discussion of trends.}
Across datasets, the largest relative improvements consistently occur on
\emph{near-OOD}, corroborating prior observations that near-OOD detection
is intrinsically more challenging than far-OOD \cite{Yang22OpenOODv1,Zhang24OpenOODv15,Yang2021SCOOD,DSAAhmed}.
Our radius-based regularization particularly benefits settings with strong
inter-class semantic similarity, where tightening ID feature manifolds
reduces spurious over-confidence on near-OOD samples.
On far-OOD, improvements are more modest but stable, as the underlying
separability is already high for most methods.

\subsection{Feature geometry analysis on CIFAR-100.}
\begin{figure}[t]
  \centering
  \includegraphics[width=0.48\linewidth]{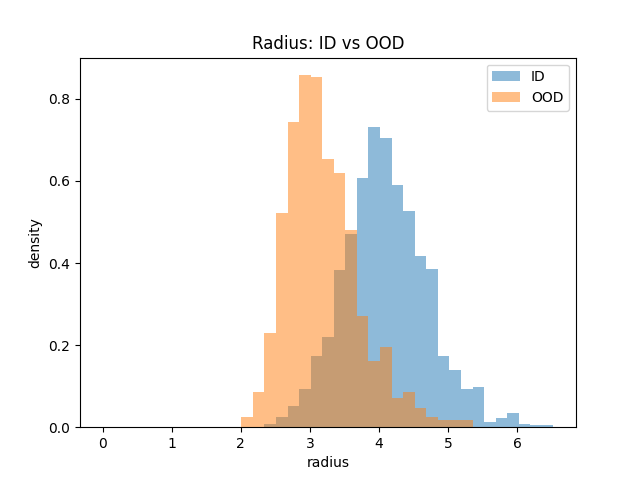}\hfill
  \includegraphics[width=0.48\linewidth]{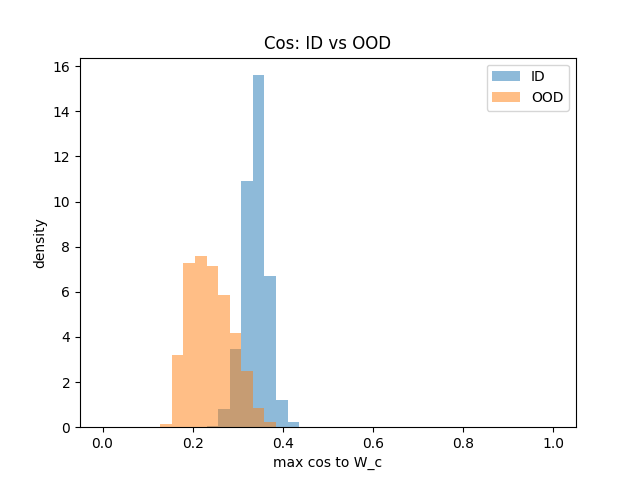}
  \caption{CIFAR-100 feature diagnostics: (left) radius $\|\mathbf{z}\|_2$; (right) max cosine to class weights $\max_c \cos(\mathbf{z},\mathbf{w}_c)$. ID (blue) vs. OOD (orange).}
  \label{fig:radius_cifar100}
\end{figure}

To inspect what \emph{BootOOD} learns, we analyze the distributions of
(i) feature \emph{radius} $\|\mathbf{z}\|_2$ and
(ii) the maximum cosine to class weights $\max_c \cos(\mathbf{z},\mathbf{w}_c)$,
computed during the \textbf{standard OpenOOD evaluation stage}.
The statistics are collected over \textbf{ID test images} and \textbf{real OOD test images}
following the OpenOOD protocol.
As shown in Fig.~\ref{fig:radius_cifar100} (left), ID features concentrate at larger radii
while OOD features concentrate at smaller radii, with only a narrow overlap;
this matches the goal of our radius self-supervision, which encourages compact,
larger-norm ID representations and suppresses large norms for OOD inputs.
In Fig.~\ref{fig:radius_cifar100} (right), ID samples exhibit consistently higher
maximal cosine to some class prototype than OOD samples, indicating that ID features
are more strongly aligned with class directions whereas OOD features remain less aligned.
Together, the two diagnostics corroborate the mechanism observed in the ablations:
the \emph{radius} terms provide the primary separability via a clear norm gap,
while the \emph{feature separation} term further reduces spurious alignment of OOD samples
with any class direction, which is particularly beneficial for near-OOD.
We observe the same qualitative pattern across seeds, indicating that the geometry induced
by \emph{BootOOD} is stable under the evaluation protocol.

\subsection{Ablations and Design Choices}
\label{sec:ablations}

We conduct ablations on \textbf{CIFAR-100} (ResNet-18), averaging results over all near-/far-OOD test sets.
We study the impact of the following design choices:
(1) the \emph{Phase-I} Neural-Collapse warm-up,
(2) the \emph{radius self-supervision} mechanism,
(3) the decomposition of radius self-supervision into \emph{classification-only} and \emph{regression-only} variants,
(4) the \emph{feature-separation} loss,
(5) the number of radius \emph{levels} $K$ used by the training-time radius head,
and (6) the strength of OOD supervision controlled by $\lambda_{\text{ood\_max}}$.
Tables~\ref{tab:ablation} and \ref{tab:cifar100_rel} report component-removal ablations against the full model, while Tables~\ref{tab:abla_k} and \ref{tab:lambda_oodmax_rel} report sensitivity to the hyperparameters $K$ and $\lambda_{\text{ood\_max}}$.

\begin{table}
  \caption{Ablation study on CIFAR-100 using ResNet-18. 
  Results show relative changes (\emph{\%}) with respect to the full BootOOD model. 
  Removing any component degrades OOD detection (lower AUROC, higher FPR@95\%).}
  \label{tab:ablation}
  \centering
  \scriptsize
  \setlength{\tabcolsep}{5pt}
  \begin{tabular}{@{}lccc@{}}
    \toprule
    Setting &
    \shortstack[c]{$\Delta$AUROC\\(\%)$\uparrow$} &
    \shortstack[c]{$\Delta$FPR@95\%\\$\downarrow$} &
    \shortstack[c]{$\Delta$ID-ACC\\(\%)$\uparrow$} \\
    \midrule
    BootOOD (full)              & ---  & ---   & ---   \\
    w/o Phase-1 warm-up         & -2.6 & +9.7  & -1.1  \\
    w/o radius self-supervision & -4.2 & +5.3  & -1.3  \\
    w/o feature separation      & -0.7 & +1.0  & -0.8  \\
    \bottomrule
  \end{tabular}
\end{table}

\paragraph{Phase-1 warm-up.}
Removing the Phase-1 warm-up leads to a clear drop in AUROC and a large increase in FPR@95\%. This shows that letting the classifier first learn stable ID prototypes before turning on OOD regularization is important for both ID accuracy and OOD detection.

\paragraph{Radius self-supervision.}
Ablating the radius self-supervision (\(\lambda_{\text{cls}}\mathcal{L}_{\text{rad-cls}}+\lambda_{\text{mse}}\mathcal{L}_{\text{rad-reg}}\)) hurts performance the most. Once these losses are removed, the model behaves close to plain CE training, and the feature norms no longer provide a strong ID vs.\ OOD separation signal.

\paragraph{Feature separation.}
Dropping the feature separation term yields a smaller but consistent degradation. Thus, separation is not the main driver but still gives a useful extra margin, especially for near-OOD close to class boundaries.

\paragraph{Effect of radius levels $K$.}
We vary the number of radius \emph{levels} $K$ used by the training-only radius head, where each level corresponds to a distinct target radius (i.e., a concentric shell). As shown in Table~\ref{tab:abla_k}, $K=1$ (degenerating to pure regression without meaningful radius classification) performs worst. Increasing $K$ initially improves near-OOD AUROC and FPR@95 by providing a richer set of discrete radial targets. Performance peaks at $K=4$ and then slightly degrades for $K=6$, suggesting that excessive radial levels introduce unnecessary complexity without clear benefits. We therefore adopt $K=4$ as the default setting for CIFAR-100 in the following experiments.

\begin{table}[t]
  \caption{Near-OOD ablation on the number of radius shells $K$ used by the training-time radius head (CIFAR-100).}
  \label{tab:abla_k}
  \centering
  \scriptsize
  \setlength{\tabcolsep}{5pt}
  \begin{tabular}{@{}lccc@{}}
    \toprule
    \textbf{$K$} 
    & $\Delta$\textbf{AUROC} (\%)$\uparrow$
    & $\Delta$\textbf{FPR@95\%} (\%)$\downarrow$
    & $\Delta$\textbf{ID-ACC} (\%)$\uparrow$ \\
    \midrule
    $K = 1$ & -1.33 & +5.14 & -1.36 \\
    $K = 3$ & -0.54 & +3.23 & -0.65 \\
    $K = 4$ & ---   & ---  & ---   \\
    $K = 6$ & -1.08 & +1.25 & -0.57 \\
    \bottomrule
  \end{tabular}
\end{table}

\paragraph{Ablation on CIFAR-100: joint vs.\ regression-only vs.\ classification-only.}
Table~\ref{tab:cifar100_rel} compares the full joint model with its single-branch variants. The classification-only variant learns discrete radial bins but lacks an absolute scale for $\lVert z-\mu\rVert$, leading to drift within bins and degraded AUROC, FPR@95, and ID accuracy. The regression-only variant enforces an absolute radius but misses the discretized margin and multi-scale gradients from the bins, further hurting detection. Combining both terms yields the best trade-off: bin-level separation from classification and radius refinement from regression, resulting in the strongest OOD separability and the smallest drops in ID-ACC.

\paragraph{Effect of $\lambda_{\text{ood\_max}}$ on CIFAR-100.} Table~\ref{tab:lambda_oodmax_rel} reports the effect of varying $\lambda_{\text{ood\_max}}$ relative to the default value $0.1$. A smaller coefficient ($0.05$) weakens the radial and directional supervision from the pseudo-OOD branch, yielding slightly worse AUROC, higher FPR@95, and a noticeable drop in ID accuracy. Increasing $\lambda_{\text{ood\_max}}$ ($0.2$ or $0.5$) makes the auxiliary loss overly dominant, over-contracting ID features and perturbing classifier directions, further degrading both AUROC and ID-ACC while increasing FPR@95. Overall, $\lambda_{\text{ood\_max}} = 0.1$ provides the best trade-off between enforcing OOD-specific supervision and preserving the geometry of ID decision regions.

\paragraph{Summary.}
Overall, all components play complementary roles.
The Phase-I warm-up is crucial for reaching a stable Neural-Collapse regime before applying OOD-specific supervision.
Radius self-supervision is the primary driver of OOD separability, and the joint formulation
that combines radial classification with regression consistently outperforms either branch alone.
Feature separation provides additional gains by reducing spurious class alignment.
A moderate number of radius levels ($K{=}4$) yields the best near-OOD performance on CIFAR-100,
while the supervision strength must be carefully balanced:
$\lambda_{\text{ood\_max}}{=}0.1$ offers the best trade-off between enforcing OOD-aware geometry
and preserving in-distribution decision structure.

\begin{table}
  \caption{Ablation on CIFAR-100: relative change (\%) w.r.t.\ the joint (full) model.}
  \label{tab:cifar100_rel}
  \centering
  \scriptsize
  \setlength{\tabcolsep}{8pt}
  \begin{tabular}{@{}lccc@{}}
    \toprule
    Setting &
    \shortstack[c]{$\Delta$AUROC\\(\%)$\uparrow$} &
    \shortstack[c]{$\Delta$FPR@95\%\\(\%)$\downarrow$} &
    \shortstack[c]{$\Delta$ID-ACC\\(\%)$\uparrow$} \\
    \midrule
    Joint (full) & --- & --- & --- \\
    
    Classification only &
    -0.65 &
    +1.84 &
    -1.92 \\
    
    Regression only &
    -0.88 &
    +4.84 &
    -0.87 \\
    \bottomrule
  \end{tabular}
\end{table}

\begin{table}
  \caption{Effect of $\lambda_{\text{ood\_max}}$ on CIFAR-100 (relative change w.r.t.\ $\lambda_{\text{ood\_max}} = 0.1$). }
  \label{tab:lambda_oodmax_rel}
  \centering
  \scriptsize
  \setlength{\tabcolsep}{8pt}
  \begin{tabular}{@{}lccc@{}}
    \toprule
    $\lambda_{\text{ood\_max}}$ &
    \shortstack[c]{$\Delta$AUROC\\(\%)$\uparrow$} &
    \shortstack[c]{$\Delta$FPR@95\%\\(\%)$\downarrow$} &
    \shortstack[c]{$\Delta$ID-ACC\\(\%)$\uparrow$} \\
    \midrule
    0.10 & ---   & ---   & ---   \\
    0.05 & -0.32 & +0.61 & -0.77 \\
    0.20 & -0.78 & +2.76 & -1.23 \\
    0.50 & -1.26 & +4.25 & -1.41 \\
    \bottomrule
  \end{tabular}
\end{table}

\section{Conclusion and Future Work}
\label{sec:conclusion}

\paragraph{Conclusion.}
We presented \textbf{BootOOD}, a \emph{training-time}, ID-only OOD detector that exploits late-phase Neural Collapse geometry to shape a simple feature-norm score. On CIFAR-10/100 and ImageNet-200 within OpenOOD---where \emph{near-OOD} is both the most challenging regime and the key leaderboard criterion---BootOOD significantly improves near-OOD AUROC and FPR95 while preserving ID top-1 accuracy. A lightweight radius head is used only during training: we synthesize pseudo-OOD features via mixup, supervise them with inner-shell radius targets, and add a directional separation penalty; the inference path remains unchanged.

\paragraph{Future Work.}
\begin{itemize}
  \item \textbf{Richer pseudo-OOD generators.} Beyond feature-space mixup, we have implemented column-element shuffle and block-shuffle generators; they tend to help far-OOD more than near-OOD and are therefore omitted from the main tables. A natural extension is to combine these generators in a unified objective or curriculum to jointly optimize near- and far-OOD performance.
  \item \textbf{Broader backbones and theory.} Applying BootOOD to larger or multimodal backbones and open-set or long-tailed settings, and further analyzing how Neural Collapse geometry aligns with radius-based scores and ID-only calibration, are promising directions.
\end{itemize}


{
    \small
    \bibliographystyle{ieeenat_fullname}
    \bibliography{main}
}
\clearpage
\setcounter{page}{1}
\maketitlesupplementary

\section{All Results}

Tables~\ref{tab:cifar10_all_short}, \ref{tab:cifar100_all_short}, and \ref{tab:imagenet200_all_short}
report the full near-/far-OOD results on CIFAR-10, CIFAR-100, and ImageNet-200, respectively.
Here, we include a comprehensive evaluation covering \emph{both} near- and far-OOD settings, as well as a wide range of post-hoc and training-time baselines.  
For each dataset, we list classical post-hoc scoring methods, recent NC-based and feature-space approaches, training-time methods with and without mixup-style OOD exposure, and all official OpenOOD baselines. 

Across all three benchmarks, our method (\textbf{BootOOD}) consistently delivers state-of-the-art or highly competitive performance in AUROC, FPR@95, AUPR-IN, and AUPR-OUT, while maintaining strong ID accuracy. The results also demonstrate that the gains hold across both near- and far-OOD
evaluation regimes, indicating that the proposed synthetic OOD exposure under Neural Collapse provides robust and transferable improvements.

\begin{table}[t]
  \captionsetup{skip=4pt}
  \caption{CIFAR-10: Near/Far-OOD performance.}
  \vspace{-2pt}
  \label{tab:cifar10_all_short}
  \centering

  \setlength{\tabcolsep}{1.6pt}
  \renewcommand{\arraystretch}{1.08}

  \resizebox{\linewidth}{!}{%
  \begin{tabular}{@{}l c c c c c@{}}
    \toprule
    \multicolumn{6}{l}{\textbf{CIFAR-10 Baseline ID Accuracy:} 95.22 ($\pm$ 0.30)} \\
    \midrule
    \textbf{Method} &
    \shortstack[c]{\textbf{AUROC}\\\textbf{(Near/Far)}} &
    \shortstack[c]{\textbf{FPR@95\%}\\\textbf{(Near/Far)}} &
    \shortstack[c]{\textbf{AUPR-IN}\\\textbf{(Near/Far)}} &
    \shortstack[c]{\textbf{AUPR-OUT}\\\textbf{(Near/Far)}} &
    \textbf{ID-ACC} \\
    \midrule

    \multicolumn{6}{l}{\textbf{\textit{Post-hoc Methods}}}\\
    Gram\cite{Sastry2020Gram} &
      52.63($\pm 5.23$)/69.74($\pm 3.61$) &
      93.52($\pm 5.33$)/69.29($\pm 2.98$) &
      54.46($\pm 3.81$)/55.30($\pm 2.66$) &
      51.10($\pm 4.58$)/79.36($\pm 3.04$) &
      \textbf{95.22}($\pm 0.30$) \\
    GradNorm\cite{Huang2021GradNorm} &
      53.77($\pm 0.98$)/58.56($\pm 3.22$) &
      95.37($\pm 0.30$)/58.56($\pm 1.12$) &
      53.18($\pm 0.44$)/36.93($\pm 0.78$) &
      55.18($\pm 2.40$)/74.20($\pm 0.34$) &
      \textbf{95.22}($\pm 0.30$) \\
    MDSEns\cite{Lee2018Mahalanobis} &
      60.46($\pm 0.30$)/74.07($\pm 0.26$) &
      92.06($\pm 0.25$)/61.09($\pm 0.33$) &
      59.37($\pm 0.19$)/63.25($\pm 0.31$) &
      59.76($\pm 0.26$)/83.37($\pm 0.28$) &
      \textbf{95.22}($\pm 0.30$) \\
    OpenGAN\cite{Kong2021OpenGAN} &
      61.45($\pm 6.54$)/58.83($\pm 17.11$) &
      77.16($\pm 11.32$)/75.39($\pm 13.07$) &
      65.56($\pm 9.81$)/45.19($\pm 11.00$) &
      57.86($\pm 12.32$)/74.46($\pm 17.44$) &
      \textbf{95.22}($\pm 0.30$) \\
    ASH\cite{Djurisic2023ASH} &
      74.11($\pm 1.06$)/78.36($\pm 2.85$) &
      89.02($\pm 1.96$)/76.65($\pm 4.33$) &
      69.37($\pm 0.85$)/55.00($\pm 2.24$) &
      75.53($\pm 1.01$)/86.68($\pm 0.30$) &
      \textbf{95.22}($\pm 0.30$) \\
    RankFeat\cite{Song2022RankFeat} &
      77.33($\pm 3.12$)/72.15($\pm 4.60$) &
      67.38($\pm 2.82$)/68.24($\pm 4.73$) &
      78.52($\pm 2.39$)/55.43($\pm 5.18$) &
      71.27($\pm 3.84$)/77.81($\pm 3.29$) &
      \textbf{95.22}($\pm 0.30$) \\
    DICE\cite{Sun2022DICE} &
      77.59($\pm 0.76$)/85.23($\pm 1.83$) &
      79.94($\pm 0.67$)/54.09($\pm 2.06$) &
      74.44($\pm 0.68$)/69.89($\pm 2.30$) &
      77.47($\pm 0.51$)/91.16($\pm 1.21$) &
      \textbf{95.22}($\pm 0.30$) \\
    KLM\cite{Hendrycks2022ScalingOOD} &
      78.80($\pm 0.90$)/82.76($\pm 0.27$) &
      86.41($\pm 0.83$)/76.41($\pm 0.11$) &
      72.04($\pm 0.87$)/54.54($\pm 0.20$) &
      80.09($\pm 0.74$)/90.24($\pm 0.07$) &
      \textbf{95.22}($\pm 0.30$) \\
    ODIN\cite{Liang2018ODIN} &
      80.25($\pm 1.55$)/87.21($\pm 0.72$) &
      84.49($\pm 0.84$)/60.90($\pm 0.88$) &
      75.11($\pm 0.67$)/69.61($\pm 0.65$) &
      81.42($\pm 0.71$)/92.55($\pm 0.74$) &
      \textbf{95.22}($\pm 0.30$) \\
    SHE\cite{Zhang2023SHE} &
      80.84($\pm 1.30$)/86.54($\pm 3.08$) &
      84.48($\pm 1.10$)/63.26($\pm 2.56$) &
      75.53($\pm 2.30$)/68.10($\pm 3.10$) &
      81.93($\pm 1.39$)/91.29($\pm 0.30$) &
      \textbf{95.22}($\pm 0.30$) \\
    OpenMax\cite{Bendale2016OpenMax} &
      86.47($\pm 0.25$)/91.02($\pm 0.24$) &
      71.57($\pm 0.23$)/42.41($\pm 0.19$) &
      86.38($\pm 0.32$)/93.61($\pm 0.22$) &
      82.91($\pm 0.15$)/77.93($\pm 0.30$) &
      \textbf{95.22}($\pm 0.30$) \\
    ReAct\cite{Sun2021ReAct} &
      86.47($\pm 0.77$)/91.02($\pm 1.83$) &
      71.57($\pm 0.98$)/42.41($\pm 0.32$) &
      82.91($\pm 0.24$)/77.93($\pm 2.07$) &
      86.38($\pm 0.69$)/93.61($\pm 1.79$) &
      \textbf{95.22}($\pm 0.30$) \\
    MDS\cite{Lee2018Mahalanobis} &
      86.72($\pm 2.58$)/90.20($\pm 1.64$) &
      46.23($\pm 1.44$)/30.31($\pm 1.33$) &
      88.06($\pm 1.92$)/82.18($\pm 2.03$) &
      83.17($\pm 1.25$)/94.22($\pm 0.89$) &
      \textbf{95.22}($\pm 0.30$) \\
    MLS\cite{Hendrycks2022ScalingOOD} &
      86.86($\pm 0.44$)/91.61($\pm 0.94$) &
      67.53($\pm 0.53$)/40.55($\pm 0.63$) &
      83.52($\pm 0.28$)/79.02($\pm 0.71$) &
      86.77($\pm 0.92$)/94.18($\pm 0.80$) &
      \textbf{95.22}($\pm 0.30$) \\
    EBO\cite{Liu20Energy}&
      86.93($\pm 0.56$)/91.74($\pm 1.21$) &
      67.54($\pm 0.44$)/40.56($\pm 0.88$) &
      83.55($\pm 0.32$)/79.12($\pm 0.96$) &
      86.94($\pm 0.47$)/94.28($\pm 0.77$) &
      \textbf{95.22}($\pm 0.30$) \\
    TempScale\cite{Guo2017Calibration} &
      87.65($\pm 0.34$)/91.27($\pm 0.55$) &
      56.84($\pm 0.32$)/33.36($\pm 0.43$) &
      85.84($\pm 0.29$)/80.81($\pm 0.51$) &
      86.07($\pm 0.22$)/93.54($\pm 0.56$) &
      \textbf{95.22}($\pm 0.30$) \\
    MSP\cite{Hendrycks2017MSP} &
      87.68($\pm 0.21$)/91.00($\pm 0.43$) &
      53.54($\pm 0.45$)/31.43($\pm 0.34$) &
      86.38($\pm 0.36$)/81.18($\pm 0.28$) &
      85.43($\pm 0.22$)/93.12($\pm 0.36$) &
      \textbf{95.22}($\pm 0.30$) \\
    VIM\cite{Wang2022ViM} &
      88.51($\pm 0.28$)/93.14($\pm 0.13$) &
      48.07($\pm 0.22$)/25.76($\pm 0.34$) &
      88.09($\pm 0.17$)/84.74($\pm 0.26$) &
      86.50($\pm 0.21$)/95.93($\pm 0.55$) &
      \textbf{95.22}($\pm 0.30$) \\
    RMDS\cite{Ren2018Mahalanobis} &
      89.53($\pm 0.35$)/92.43($\pm 0.41$) &
      42.19($\pm 0.31$)/24.38($\pm 0.33$) &
      89.79($\pm 0.26$)/84.90($\pm 0.36$) &
      87.48($\pm 0.34$)/94.11($\pm 0.44$) &
      \textbf{95.22}($\pm 0.30$) \\
    KNN\cite{Sun2022DeepKNN} &
      90.70($\pm 0.10$)/93.11($\pm 0.17$) &
      34.54($\pm 0.22$)/23.88($\pm 0.33$) &
      91.73($\pm 0.21$)/87.26($\pm 0.18$) &
      88.71($\pm 0.36$)/94.92($\pm 0.20$) &
      \textbf{95.22}($\pm 0.30$) \\
    NECO\cite{Ammar2024NECO} &
      91.52($\pm 2.63$)/95.32($\pm 5.32$) &
      37.32($\pm 0.61$)/\textbf{19.86}($\pm 1.98$) &
      91.86($\pm 1.42$)/89.59($\pm 2.57$) &
      \textbf{90.23}($\pm 1.01$)/96.57($\pm 5.56$) &
      \textbf{95.22}($\pm 0.30$) \\
    \textbf{BootOOD (Ours)} &
      \textbf{92.40}($\pm 0.28$)/\textbf{96.31}($\pm 0.41$) &
      \textbf{31.34}($\pm 0.27$)/33.13($\pm 0.33$) &
      \textbf{93.75}($\pm 0.36$)/\textbf{93.87}($\pm 0.51$) &
      89.94($\pm 0.35$)/\textbf{96.93}($\pm 0.26$) &
      95.08($\pm 0.43$) \\
    \midrule

    \multicolumn{6}{l}{\textbf{\textit{Training Methods w/o Outlier Data}}} \\
    LogitNorm\cite{Wei2022LogitNorm} &
      65.64($\pm 0.05$)/77.70($\pm 0.07$) &
      79.18($\pm 0.15$)/62.01($\pm 0.16$) &
      68.99($\pm 1.08$)/63.73($\pm 0.09$) &
      63.00($\pm 0.99$)/88.68($\pm 0.08$) &
      59.56($\pm 0.25$) \\
    MOS\cite{Huang2021MOS} &
      69.16($\pm 2.96$)/61.84($\pm 5.83$) &
      86.32($\pm 3.32$)/71.46($\pm 4.67$) &
      66.06($\pm 1.66$)/50.79($\pm 4.20$) &
      68.27($\pm 2.78$)/82.22($\pm 7.09$) &
      \textbf{95.16}($\pm 0.36$) \\
    RotPred\cite{Hendrycks2019SSL} &
      86.62($\pm 0.16$)/94.70($\pm 0.30$) &
      56.12($\pm 0.26$)/33.04($\pm 0.30$) &
      83.27($\pm 0.27$)/81.50($\pm 0.30$) &
      77.61($\pm 0.23$)/92.53($\pm 0.30$) &
      80.19($\pm 0.49$) \\
    VOS\cite{Du2022VOS} &
      86.72($\pm 0.44$)/90.82($\pm 0.90$) &
      67.31($\pm 0.47$)/45.37($\pm 0.57$) &
      83.89($\pm 0.39$)/73.44($\pm 0.78$) &
      86.52($\pm 0.46$)/94.28($\pm 0.94$) &
      95.00($\pm 0.58$) \\
    ARPL\cite{Chen2022ARPL} &
      87.29($\pm 0.25$)/89.49($\pm 0.34$) &
      41.26($\pm 0.16$)/32.49($\pm 0.29$) &
      89.17($\pm 0.19$)/82.61($\pm 0.21$) &
      83.11($\pm 0.31$)/91.74($\pm 0.14$) &
      93.62($\pm 0.09$) \\
    NPOS\cite{Tao2023NPOS} &
      87.31($\pm 0.31$)/91.83($\pm 0.47$) &
      45.02($\pm 0.42$)/33.13($\pm 0.33$) &
      86.82($\pm 0.36$)/80.34($\pm 0.34$) &
      76.49($\pm 0.27$)/91.12($\pm 0.42$) &
      N/A \\
    G-ODIN\cite{Hsu2020GODIN} &
      89.39($\pm 0.59$)/95.72($\pm 0.49$) &
      43.75($\pm 0.54$)/20.38($\pm 0.58$) &
      89.74($\pm 0.48$)/90.84($\pm 0.52$) &
      88.46($\pm 0.68$)/97.44($\pm 0.60$) &
      94.80($\pm 0.25$) \\
    ConfBranch\cite{DeVries2018Confidence}&
      89.84($\pm 0.24$)/93.69($\pm 0.30$) &
      31.97($\pm 0.13$)/20.65($\pm 0.16$) &
      91.72($\pm 0.29$)/89.83($\pm 0.33$) &
      85.84($\pm 0.54$)/94.75($\pm 0.47$) &
      94.24($\pm 0.11$) \\
    CIDER\cite{Ming2023CIDER} &
      90.26($\pm 0.20$)/94.83($\pm 0.36$) &
      32.94($\pm 0.34$)/\textbf{20.00}($\pm 0.31$) &
      91.88($\pm 0.18$)/89.57($\pm 0.35$) &
      87.72($\pm 0.27$)/96.24($\pm 0.26$) &
      N/A \\
    \textbf{BootOOD (Ours)} &
      \textbf{92.40}($\pm 0.28$)/\textbf{96.31}($\pm 0.41$) &
      \textbf{31.34}($\pm 0.27$)/33.13($\pm 0.33$) &
      \textbf{93.75}($\pm 0.36$)/\textbf{93.87}($\pm 0.51$) &
      \textbf{89.94}($\pm 0.35$)/\textbf{96.93}($\pm 0.26$) &
      95.08($\pm 0.43$) \\
    \midrule

    \multicolumn{6}{l}{\textbf{\textit{Training Methods w/ Outlier Data}}} \\
    MixOE\cite{Zhang2023MixOE}&
      88.57($\pm 0.88$)/92.35($\pm 0.75$) &
      56.07($\pm 0.74$)/31.44($\pm 0.66$) &
      86.43($\pm 0.86$)/78.53($\pm 0.59$) &
      87.61($\pm 0.69$)/95.23($\pm 0.83$) &
      94.56($\pm 0.34$) \\
    PFS\cite{Wu2025SepLoss} &
      89.10($\pm 1.25$)/92.96($\pm 2.29$) &
      37.38($\pm 0.83$)/12.21($\pm 2.21$) &
      89.67($\pm 0.96$)/88.59($\pm 0.87$) &
      81.01($\pm 1.04$)/91.14($\pm 1.59$) &
      86.06($\pm 0.54$) \\
    MCD\cite{Yu2019MCD} &
      89.21($\pm 0.14$)/93.24($\pm 1.23$) &
      25.65($\pm 0.21$)/24.30($\pm 2.30$) &
      88.80($\pm 0.18$)/88.40($\pm 1.44$) &
      91.21($\pm 1.15$)/89.00($\pm 0.99$) &
      93.67($\pm 0.04$) \\
    UDG\cite{Yang2021SCOOD} &
      89.77($\pm 0.24$)/93.24($\pm 0.93$) &
      36.18($\pm 0.26$)/21.27($\pm 1.13$) &
      90.65($\pm 0.19$)/88.40($\pm 1.43$) &
      87.05($\pm 0.17$)/94.06($\pm 0.93$) &
      93.77($\pm 0.87$) \\
    OE\cite{Hendrycks2019OE} &
      \textbf{94.14}($\pm 0.21$)/95.73($\pm 0.14$) &
      \textbf{19.65}($\pm 0.33$)/\textbf{11.39}($\pm 0.19$) &
      \textbf{94.57}($\pm 0.28$)/93.23($\pm 0.11$) &
      \textbf{93.88}($\pm 0.19$)/\textbf{98.31}($\pm 0.28$) &
      94.24($\pm 0.24$) \\
    \textbf{BootOOD (Ours)} &
      92.40($\pm 0.28$)/\textbf{96.31}($\pm 0.41$) &
      31.34($\pm 0.27$)/33.13($\pm 0.33$) &
      93.75($\pm 0.36$)/\textbf{93.87}($\pm 0.51$) &
      89.94($\pm 0.35$)/96.93($\pm 0.26$) &
      \textbf{95.08}($\pm 0.43$) \\

    \bottomrule
  \end{tabular}
  }
\end{table}

\begin{table}[t]
  \captionsetup{skip=4pt}
  \caption{CIFAR-100: Near/Far-OOD performance.}
  \vspace{-2pt}
  \label{tab:cifar100_all_short}
  \centering

  \setlength{\tabcolsep}{1.6pt}
  \renewcommand{\arraystretch}{1.08}

  \resizebox{\linewidth}{!}{%
  \begin{tabular}{@{}l c c c c c@{}}
    \toprule
    \multicolumn{6}{l}{\textbf{CIFAR-100 Baseline ID Accuracy:} 77.19 ($\pm$ 0.13)} \\
    \midrule
    \textbf{Method} &
    \shortstack[c]{\textbf{AUROC}\\\textbf{(Near/Far)}} &
    \shortstack[c]{\textbf{FPR@95\%}\\\textbf{(Near/Far)}} &
    \shortstack[c]{\textbf{AUPR-IN}\\\textbf{(Near/Far)}} &
    \shortstack[c]{\textbf{AUPR-OUT}\\\textbf{(Near/Far)}} &
    \textbf{ID-ACC} \\
    \midrule

    \multicolumn{6}{l}{\textbf{\textit{Post-hoc Methods}}} \\
    MDSEns\cite{Lee2018Mahalanobis}  &
      46.71($\pm 0.24$)/66.45($\pm 0.36$) &
      95.84($\pm 0.24$)/66.97($\pm 0.69$) &
      50.04($\pm 0.22$)/57.34($\pm 0.48$) &
      44.05($\pm 0.22$)/76.94($\pm 0.56$) &
      77.19($\pm 0.13$) \\
    NECO\cite{Ammar2024NECO} &
      47.91($\pm 0.50$)/64.82($\pm 0.70$) &
      95.57($\pm 0.20$)/73.68($\pm 0.35$) &
      50.94($\pm 0.35$)/51.89($\pm 0.48$) &
      45.70($\pm 0.32$)/76.98($\pm 0.55$) &
      77.19($\pm 0.13$) \\
    Gram\cite{Sastry2020Gram} &
      51.22($\pm 0.55$)/74.59($\pm 0.90$) &
      92.48($\pm 0.77$)/63.09($\pm 1.08$) &
      55.08($\pm 0.52$)/61.23($\pm 0.80$) &
      46.83($\pm 0.58$)/84.28($\pm 0.88$) &
      77.19($\pm 0.13$) \\
    MDS\cite{Lee2018Mahalanobis}  &
      59.15($\pm 0.22$)/69.44($\pm 0.40$) &
      82.76($\pm 0.09$)/70.46($\pm 1.39$) &
      64.23($\pm 0.20$)/55.49($\pm 0.85$) &
      51.56($\pm 0.20$)/81.68($\pm 0.95$) &
      77.19($\pm 0.13$) \\
    RankFeat\cite{Song2022RankFeat} &
      63.07($\pm 1.00$)/68.31($\pm 1.20$) &
      79.96($\pm 1.28$)/68.89($\pm 1.42$) &
      67.92($\pm 0.95$)/54.91($\pm 1.00$) &
      54.56($\pm 1.05$)/78.57($\pm 1.10$) &
      77.19($\pm 0.13$) \\
    GradNorm\cite{Huang2021GradNorm} &
      69.38($\pm 0.45$)/68.25($\pm 0.70$) &
      86.12($\pm 0.47$)/82.79($\pm 1.05$) &
      69.08($\pm 0.40$)/46.28($\pm 0.75$) &
      66.84($\pm 0.44$)/82.02($\pm 0.80$) &
      77.19($\pm 0.13$) \\
    VIM\cite{Wang2022ViM} &
      74.29($\pm 0.35$)/82.66($\pm 0.85$) &
      62.95($\pm 0.13$)/\textbf{49.73}($\pm 0.62$) &
      78.77($\pm 0.26$)/71.31($\pm 0.58$) &
      67.19($\pm 0.28$)/89.31($\pm 0.66$) &
      77.19($\pm 0.13$) \\
    OpenGAN\cite{Kong2021OpenGAN} &
      75.93($\pm 0.70$)/71.86($\pm 1.80$) &
      68.78($\pm 1.26$)/67.40($\pm 7.16$) &
      72.82($\pm 0.90$)/57.02($\pm 1.50$) &
      61.20($\pm 0.95$)/79.82($\pm 1.60$) &
      77.19($\pm 0.13$) \\
    OpenMax\cite{Bendale2016OpenMax} &
      75.98($\pm 0.30$)/78.85($\pm 0.35$) &
      55.58($\pm 0.25$)/54.77($\pm 0.41$) &
      82.04($\pm 0.28$)/66.81($\pm 0.40$) &
      64.62($\pm 0.32$)/83.76($\pm 0.45$) &
      77.19($\pm 0.13$) \\
    KLM\cite{Hendrycks2022ScalingOOD} &
      77.41($\pm 0.28$)/75.63($\pm 0.36$) &
      79.48($\pm 0.25$)/70.16($\pm 0.52$) &
      75.67($\pm 0.26$)/57.52($\pm 0.40$) &
      71.69($\pm 0.28$)/85.65($\pm 0.45$) &
      77.19($\pm 0.13$) \\
    SHE\cite{Zhang2023SHE} &
      78.24($\pm 0.28$)/77.81($\pm 0.38$) &
      59.30($\pm 0.18$)/62.74($\pm 1.16$) &
      82.15($\pm 0.26$)/62.94($\pm 0.33$) &
      72.04($\pm 0.30$)/84.82($\pm 0.40$) &
      77.19($\pm 0.13$) \\
    ASH\cite{Djurisic2023ASH}&
      78.61($\pm 0.24$)/79.28($\pm 0.66$) &
      66.05($\pm 0.15$)/62.64($\pm 2.46$) &
      80.34($\pm 0.15$)/62.77($\pm 1.55$) &
      72.71($\pm 0.17$)/86.72($\pm 0.44$) &
      77.19($\pm 0.13$) \\
    ODIN\cite{Hsu2020GODIN} &
      79.22($\pm 0.26$)/78.89($\pm 0.31$) &
      58.47($\pm 0.11$)/57.74($\pm 0.21$) &
      82.76($\pm 0.21$)/66.37($\pm 0.33$) &
      73.46($\pm 0.24$)/85.89($\pm 0.37$) &
      77.19($\pm 0.13$) \\
    DICE\cite{Sun2022DICE} &
      79.63($\pm 0.28$)/79.49($\pm 0.34$) &
      58.10($\pm 0.23$)/55.95($\pm 0.18$) &
      82.53($\pm 0.24$)/67.22($\pm 0.31$) &
      72.05($\pm 0.28$)/86.15($\pm 0.38$) &
      77.19($\pm 0.13$) \\
    KNN\cite{Sun2022DeepKNN} &
      80.02($\pm 0.27$)/\textbf{82.79}($\pm 0.34$) &
      61.31($\pm 0.15$)/54.04($\pm 0.17$) &
      81.29($\pm 0.22$)/69.33($\pm 0.30$) &
      74.86($\pm 0.25$)/89.38($\pm 0.36$) &
      77.19($\pm 0.13$) \\
    MSP\cite{Hendrycks2017MSP} &
      80.11($\pm 0.28$)/78.01($\pm 0.32$) &
      54.75($\pm 0.11$)/59.08($\pm 0.44$) &
      83.72($\pm 0.22$)/64.28($\pm 0.38$) &
      74.26($\pm 0.24$)/84.81($\pm 0.42$) &
      77.19($\pm 0.13$) \\
    RMDS\cite{Ren2018Mahalanobis} &
      80.69($\pm 0.29$)/81.98($\pm 0.34$) &
      91.13($\pm 0.11$)/85.38($\pm 0.42$) &
      58.19($\pm 0.24$)/41.73($\pm 0.40$) &
      55.96($\pm 0.26$)/76.53($\pm 0.44$) &
      77.19($\pm 0.13$) \\
    ReAct\cite{Sun2021ReAct} &
      80.93($\pm 0.24$)/80.15($\pm 0.30$) &
      56.76($\pm 0.05$)/56.29($\pm 0.49$) &
      83.47($\pm 0.18$)/66.05($\pm 0.35$) &
      74.44($\pm 0.22$)/86.08($\pm 0.40$) &
      77.19($\pm 0.13$) \\
    EBO\cite{Liu20Energy} &
      81.12($\pm 0.26$)/80.03($\pm 0.31$) &
      55.52($\pm 0.08$)/56.41($\pm 0.61$) &
      83.77($\pm 0.22$)/66.78($\pm 0.40$) &
      74.49($\pm 0.24$)/85.82($\pm 0.46$) &
      77.19($\pm 0.13$) \\
    TempScale\cite{Guo2017Calibration} &
      81.46($\pm 0.27$)/78.93($\pm 0.33$) &
      54.78($\pm 0.07$)/58.24($\pm 0.51$) &
      84.03($\pm 0.20$)/65.16($\pm 0.42$) &
      74.87($\pm 0.22$)/85.31($\pm 0.46$) &
      77.19($\pm 0.13$) \\
    MLS\cite{Hendrycks2022ScalingOOD} &
      81.56($\pm 0.23$)/80.12($\pm 0.33$) &
      55.48($\pm 0.07$)/56.54($\pm 0.57$) &
      83.88($\pm 0.18$)/\textbf{79.60}($\pm 0.38$) &
      74.93($\pm 0.22$)/66.65($\pm 0.42$) &
      77.19($\pm 0.13$) \\
    \textbf{BootOOD (Ours)} &
      \textbf{83.09}($\pm 0.23$)/81.99($\pm 0.34$) &
      \textbf{51.03}($\pm 0.36$)/53.33($\pm 0.62$) &
      \textbf{84.93}($\pm 0.24$)/69.74($\pm 0.57$) &
      \textbf{77.65}($\pm 0.28$)/\textbf{87.38}($\pm 0.29$) &
      \textbf{78.89}($\pm 0.12$) \\
    \midrule

    \multicolumn{6}{l}{\textbf{\textit{Training Methods w/o Outlier Data}}} \\
    ConfBranch\cite{DeVries2018Confidence} &
      71.12($\pm 0.62$)/68.09($\pm 1.83$) &
      70.96($\pm 0.90$)/72.05($\pm 1.30$) &
      75.05($\pm 0.55$)/53.90($\pm 0.95$) &
      62.71($\pm 0.60$)/76.54($\pm 1.20$) &
      75.82($\pm 0.27$) \\
    CIDER\cite{Ming2023CIDER} &
      74.71($\pm 0.39$)/77.12($\pm 0.68$) &
      69.96($\pm 0.28$)/59.22($\pm 0.50$) &
      76.38($\pm 0.25$)/65.40($\pm 0.38$) &
      66.85($\pm 0.30$)/84.17($\pm 0.55$) &
      N/A \\
    ARPL\cite{Chen2022ARPL} &
      75.11($\pm 0.93$)/71.09($\pm 1.80$) &
      63.12($\pm 0.70$)/67.45($\pm 1.20$) &
      79.21($\pm 0.60$)/57.77($\pm 0.95$) &
      67.11($\pm 0.65$)/80.32($\pm 1.10$) &
      69.72($\pm 1.08$) \\
    RotPred\cite{Hendrycks2019SSL} &
      75.98($\pm 0.16$)/\textbf{87.96}($\pm 0.13$) &
      63.60($\pm 0.25$)/34.24($\pm 0.22$) &
      79.35($\pm 0.18$)/81.87($\pm 0.20$) &
      68.71($\pm 0.20$)/91.16($\pm 0.22$) &
      75.15($\pm 0.38$) \\
    NPOS\cite{Tao2023NPOS} &
      76.41($\pm 0.37$)/80.98($\pm 1.55$) &
      67.11($\pm 0.26$)/66.67($\pm 1.05$) &
      78.21($\pm 0.24$)/57.76($\pm 0.70$) &
      66.72($\pm 0.28$)/84.13($\pm 1.00$) &
      N/A \\
    G-ODIN\cite{Hsu2020GODIN} &
      77.68($\pm 0.28$)/85.91($\pm 1.58$) &
      67.92($\pm 0.40$)/\textbf{43.39}($\pm 1.10$) &
      78.25($\pm 0.30$)/76.10($\pm 0.85$) &
      71.40($\pm 0.35$)/\textbf{91.55}($\pm 1.20$) &
      74.69($\pm 0.04$) \\
    LogitNorm\cite{Wei2022LogitNorm} &
      78.46($\pm 0.31$)/81.49($\pm 1.26$) &
      63.43($\pm 0.22$)/54.59($\pm 0.85$) &
      80.03($\pm 0.20$)/72.04($\pm 0.60$) &
      72.71($\pm 0.24$)/84.17($\pm 0.90$) &
      75.93($\pm 0.17$) \\
    MOS\cite{Huang2021MOS}&
      80.74($\pm 0.18$)/79.33($\pm 1.21$) &
      55.06($\pm 0.20$)/61.20($\pm 0.80$) &
      83.39($\pm 0.18$)/63.22($\pm 0.65$) &
      73.87($\pm 0.20$)/85.57($\pm 0.90$) &
      76.62($\pm 0.20$) \\
    VOS\cite{Du2022VOS} &
      81.02($\pm 0.29$)/79.85($\pm 0.09$) &
      54.99($\pm 0.18$)/55.67($\pm 0.08$) &
      83.81($\pm 0.22$)/68.08($\pm 0.15$) &
      74.19($\pm 0.24$)/86.14($\pm 0.18$) &
      77.09($\pm 0.10$) \\
    \textbf{BootOOD (Ours)} &
      \textbf{83.09}($\pm 0.23$)/81.99($\pm 0.34$) &
      \textbf{51.03}($\pm 0.36$)/53.33($\pm 0.62$) &
      \textbf{84.93}($\pm 0.24$)/\textbf{69.74}($\pm 0.57$) &
      \textbf{77.65}($\pm 0.28$)/87.38($\pm 0.29$) &
      \textbf{78.89}($\pm 0.12$) \\
    \midrule

    \multicolumn{6}{l}{\textbf{\textit{Training Methods w/ Outlier Data}}} \\
    PFS\cite{Wu2025SepLoss} &
      72.19($\pm 0.40$)/67.89($\pm 0.85$) &
      61.59($\pm 0.30$)/64.80($\pm 0.60$) &
      77.04($\pm 0.35$)/53.62($\pm 0.55$) &
      62.12($\pm 0.40$)/75.56($\pm 0.70$) &
      54.39($\pm 0.42$) \\
    UDG\cite{Yang2021SCOOD} &
      77.63($\pm 0.20$)/79.88($\pm 1.77$) &
      60.39($\pm 0.22$)/60.38($\pm 0.95$) &
      81.06($\pm 0.18$)/67.40($\pm 0.55$) &
      71.22($\pm 0.24$)/85.27($\pm 0.70$) &
      71.93($\pm 0.64$) \\
    MCD\cite{Yu2019MCD} &
      79.72($\pm 0.25$)/73.18($\pm 0.78$) &
      57.81($\pm 0.30$)/65.43($\pm 0.70$) &
      82.05($\pm 0.22$)/56.11($\pm 0.45$) &
      73.11($\pm 0.24$)/83.25($\pm 0.60$) &
      74.98($\pm 0.38$) \\
    MixOE\cite{Zhang2023MixOE} &
      80.88($\pm 0.20$)/74.67($\pm 1.44$) &
      55.04($\pm 0.18$)/67.10($\pm 0.75$) &
      83.84($\pm 0.16$)/57.80($\pm 0.52$) &
      74.89($\pm 0.20$)/84.07($\pm 0.62$) &
      75.30($\pm 0.42$) \\
    OE\cite{Hendrycks2019OE} &
      \textbf{85.88}($\pm 0.20$)/79.67($\pm 1.49$) &
      55.04($\pm 0.18$)/67.10($\pm 0.85$) &
      83.84($\pm 0.16$)/57.80($\pm 0.55$) &
      74.89($\pm 0.20$)/84.07($\pm 0.66$) &
      75.30($\pm 0.42$) \\
    \textbf{BootOOD (Ours)} &
       83.09($\pm 0.23$)/\textbf{81.99}($\pm 0.34$) &
       \textbf{51.03}($\pm 0.36$)/\textbf{53.33}($\pm 0.62$) &
       \textbf{84.93}($\pm 0.24$)/\textbf{69.74}($\pm 0.57$) &
       \textbf{77.65}($\pm 0.28$)/\textbf{87.38}($\pm 0.29$) &
       \textbf{78.89}($\pm 0.12$) \\

    \bottomrule
  \end{tabular}
  }
\end{table}

\begin{table}[t]
  \captionsetup{skip=4pt}
  \caption{ImageNet-200: Near/Far-OOD performance.}
  \vspace{-2pt}
  \label{tab:imagenet200_all_short}
  \centering

  \setlength{\tabcolsep}{1.6pt}
  \renewcommand{\arraystretch}{1.08}

  \resizebox{\linewidth}{!}{%
  \begin{tabular}{@{}l c c c c c@{}}
    \toprule
    \multicolumn{6}{l}{\textbf{ImageNet-200 Baseline ID Accuracy:} 86.37 ($\pm$ 0.09)} \\
    \midrule
    \textbf{Method} &
    \shortstack[c]{\textbf{AUROC}\\\textbf{(Near/Far)}} &
    \shortstack[c]{\textbf{FPR@95\%}\\\textbf{(Near/Far)}} &
    \shortstack[c]{\textbf{AUPR-IN}\\\textbf{(Near/Far)}} &
    \shortstack[c]{\textbf{AUPR-OUT}\\\textbf{(Near/Far)}} &
    \textbf{ID-ACC} \\
    \midrule

    \multicolumn{6}{l}{\textbf{\textit{Post-hoc Methods}}} \\
    RankFeat\cite{Song2022RankFeat} &
      50.99($\pm 1.25$)/53.93($\pm 1.33$) &
      91.83($\pm 1.85$)/87.17($\pm 1.60$) &
      \textbf{71.43}($\pm 1.20$)/83.57($\pm 1.25$) &
      33.71($\pm 1.50$)/22.01($\pm 1.40$) &
      86.37($\pm 0.09$) \\
    MDSEns\cite{Lee2018Mahalanobis} &
      54.32($\pm 1.20$)/69.27($\pm 0.95$) &
      91.75($\pm 0.55$)/80.96($\pm 0.64$) &
      42.16($\pm 0.38$)/66.06($\pm 0.52$) &
      64.81($\pm 0.45$)/69.62($\pm 0.60$) &
      86.37($\pm 0.09$) \\
    NECO\cite{Ammar2024NECO} &
      55.89($\pm 0.92$)/81.38($\pm 0.64$) &
      86.98($\pm 0.58$)/57.68($\pm 0.65$) &
      45.81($\pm 0.35$)/61.53($\pm 0.40$) &
      63.92($\pm 0.42$)/54.40($\pm 0.45$) &
      86.39($\pm 0.09$) \\
    OpenGAN\cite{Kong2021OpenGAN} &
      59.79($\pm 1.35$)/73.18($\pm 1.12$) &
      73.15($\pm 4.07$)/68.43($\pm 3.85$) &
      61.92($\pm 3.50$)/69.15($\pm 3.20$) &
      67.80($\pm 3.70$)/74.28($\pm 3.40$) &
      86.37($\pm 0.09$) \\
    MDS\cite{Lee2018Mahalanobis} &
      61.93($\pm 1.02$)/74.72($\pm 0.82$) &
      79.11($\pm 0.85$)/61.66($\pm 0.60$) &
      50.93($\pm 0.48$)/76.99($\pm 0.55$) &
      67.68($\pm 0.50$)/70.80($\pm 0.58$) &
      86.37($\pm 0.09$) \\
    Gram\cite{Sastry2020Gram} &
      67.67($\pm 0.96$)/71.19($\pm 0.75$) &
      86.40($\pm 1.25$)/84.36($\pm 0.50$) &
      50.58($\pm 0.80$)/65.73($\pm 0.55$) &
      75.63($\pm 0.95$)/72.75($\pm 0.58$) &
      86.37($\pm 0.09$) \\
    GradNorm\cite{Huang2021GradNorm} &
      72.75($\pm 0.84$)/84.26($\pm 0.71$) &
      82.67($\pm 0.60$)/66.45($\pm 0.75$) &
      54.77($\pm 0.45$)/80.09($\pm 0.58$) &
      80.19($\pm 0.52$)/86.54($\pm 0.66$) &
      86.37($\pm 0.09$) \\
    VIM\cite{Wang2022ViM} &
      78.68($\pm 0.62$)/91.26($\pm 0.41$) &
      59.19($\pm 0.28$)/27.20($\pm 0.22$) &
      64.60($\pm 0.22$)/91.92($\pm 0.20$) &
      81.61($\pm 0.25$)/90.01($\pm 0.22$) &
      86.37($\pm 0.09$) \\
    SHE\cite{Zhang2023SHE} &
      80.18($\pm 0.53$)/89.81($\pm 0.38$) &
      66.80($\pm 0.32$)/42.17($\pm 0.38$) &
      64.25($\pm 0.26$)/89.10($\pm 0.28$) &
      84.20($\pm 0.24$)/90.05($\pm 0.30$) &
      86.37($\pm 0.09$) \\
    OpenMax\cite{Bendale2016OpenMax} &
      80.27($\pm 0.42$)/90.20($\pm 0.31$) &
      63.48($\pm 0.35$)/33.12($\pm 0.25$) &
      64.71($\pm 0.28$)/90.93($\pm 0.26$) &
      81.42($\pm 0.22$)/85.13($\pm 0.24$) &
      86.37($\pm 0.09$) \\
    ODIN\cite{Liang2018ODIN} &
      80.27($\pm 0.47$)/91.71($\pm 0.34$) &
      66.76($\pm 0.33$)/34.23($\pm 0.26$) &
      63.92($\pm 0.26$)/91.56($\pm 0.28$) &
      85.02($\pm 0.23$)/91.29($\pm 0.24$) &
      86.37($\pm 0.09$) \\
    KLM\cite{Hendrycks2022ScalingOOD} &
      80.76($\pm 0.58$)/88.53($\pm 0.45$) &
      70.26($\pm 0.33$)/40.90($\pm 0.30$) &
      62.99($\pm 0.30$)/88.26($\pm 0.25$) &
      83.41($\pm 0.28$)/84.22($\pm 0.27$) &
      86.37($\pm 0.09$) \\
    KNN\cite{Sun2022DeepKNN} &
      81.58($\pm 0.52$)/93.16($\pm 0.39$) &
      60.18($\pm 0.20$)/27.27($\pm 0.21$) &
      64.82($\pm 0.18$)/92.73($\pm 0.20$) &
      85.72($\pm 0.22$)/93.48($\pm 0.22$) &
      86.37($\pm 0.09$) \\
    ReAct\cite{Sun2021ReAct} &
      81.87($\pm 0.46$)/92.31($\pm 0.27$) &
      62.49($\pm 0.32$)/28.50($\pm 0.25$) &
      65.85($\pm 0.28$)/92.53($\pm 0.22$) &
      85.38($\pm 0.25$)/91.31($\pm 0.24$) &
      86.37($\pm 0.09$) \\
    ASH\cite{Djurisic2023ASH} &
      82.38($\pm 0.40$)/93.90($\pm 0.26$) &
      64.89($\pm 0.25$)/27.29($\pm 0.23$) &
      65.53($\pm 0.20$)/93.07($\pm 0.20$) &
      87.03($\pm 0.22$)/\textbf{94.15}($\pm 0.24$) &
      86.37($\pm 0.09$) \\
    EBO\cite{Liu20Energy} &
      82.50($\pm 0.37$)/90.86($\pm 0.29$) &
      60.24($\pm 0.22$)/34.86($\pm 0.28$) &
      66.93($\pm 0.25$)/90.99($\pm 0.22$) &
      85.48($\pm 0.20$)/89.85($\pm 0.25$) &
      86.37($\pm 0.09$) \\
    RMDS\cite{Ren2018Mahalanobis} &
      82.57($\pm 0.46$)/88.06($\pm 0.52$) &
      65.94($\pm 0.38$)/43.08($\pm 0.40$) &
      69.85($\pm 0.33$)/88.29($\pm 0.28$) &
      83.72($\pm 0.30$)/86.94($\pm 0.32$) &
      86.37($\pm 0.09$) \\
    MLS\cite{Hendrycks2022ScalingOOD} &
      82.90($\pm 0.40$)/91.11($\pm 0.36$) &
      59.76($\pm 0.18$)/34.03($\pm 0.25$) &
      67.19($\pm 0.22$)/91.26($\pm 0.19$) &
      85.96($\pm 0.20$)/90.10($\pm 0.22$) &
      86.37($\pm 0.09$) \\
    TempScale\cite{Guo2017Calibration} &
      83.24($\pm 0.33$)/\textbf{93.98}($\pm 0.28$) &
      57.29($\pm 0.32$)/\textbf{26.46}($\pm 0.20$) &
      69.04($\pm 0.27$)/\textbf{93.54}($\pm 0.24$) &
      \textbf{88.38}($\pm 0.25$)/94.05($\pm 0.22$) &
      86.37($\pm 0.09$) \\
    MSP\cite{Hendrycks2017MSP} &
      83.34($\pm 0.38$)/90.13($\pm 0.36$) &
      54.82($\pm 0.30$)/35.43($\pm 0.28$) &
      68.96($\pm 0.24$)/90.66($\pm 0.25$) &
      85.95($\pm 0.25$)/88.71($\pm 0.22$) &
      86.37($\pm 0.09$) \\
    \textbf{BootOOD (Ours)} &
      \textbf{84.16}($\pm 0.28$)/89.32($\pm 0.62$) &
      \textbf{53.16}($\pm 0.22$)/33.13($\pm 0.49$) &
      68.17($\pm 0.30$)/89.78($\pm 0.28$) &
      86.78($\pm 0.65$)/87.59($\pm 0.14$) &
      \textbf{86.60}($\pm 0.15$) \\
    \midrule

    \multicolumn{6}{l}{\textbf{\textit{Training Methods w/o Outlier Data}}} \\
    VOS\cite{Du2022VOS} &
      66.17($\pm 0.52$)/75.80($\pm 1.04$) &
      81.24($\pm 1.05$)/67.99($\pm 1.10$) &
      52.10($\pm 0.60$)/75.52($\pm 0.78$) &
      72.51($\pm 0.66$)/75.67($\pm 0.74$) &
      49.67($\pm 0.30$) \\
    MOS\cite{Huang2021MOS} &
      67.56($\pm 0.58$)/79.30($\pm 1.05$) &
      74.42($\pm 0.68$)/50.90($\pm 0.88$) &
      55.99($\pm 0.40$)/82.33($\pm 0.74$) &
      71.65($\pm 0.46$)/71.20($\pm 0.82$) &
      85.18($\pm 0.24$) \\
    G-ODIN\cite{Hsu2020GODIN} &
      76.86($\pm 0.22$)/91.85($\pm 0.19$) &
      70.84($\pm 0.36$)/31.09($\pm 0.24$) &
      59.89($\pm 0.27$)/91.81($\pm 0.23$) &
      83.48($\pm 0.31$)/91.50($\pm 0.26$) &
      85.12($\pm 0.20$) \\
    NPOS\cite{Tao2023NPOS} &
      79.25($\pm 0.45$)/\textbf{94.38}($\pm 0.10$) &
      63.60($\pm 0.34$)/27.25($\pm 0.24$) &
      65.00($\pm 0.30$)/\textbf{93.88}($\pm 0.22$) &
      83.30($\pm 0.28$)/\textbf{93.55}($\pm 0.24$) &
      N/A \\
    ConfBranch\cite{DeVries2018Confidence} &
      79.72($\pm 0.29$)/91.18($\pm 0.27$) &
      60.23($\pm 0.34$)/32.57($\pm 0.29$) &
      65.15($\pm 0.27$)/91.35($\pm 0.24$) &
      83.24($\pm 0.30$)/89.91($\pm 0.28$) &
      86.14($\pm 0.12$) \\
    CIDER\cite{Ming2023CIDER} &
      80.72($\pm 1.38$)/90.52($\pm 1.44$) &
      61.40($\pm 0.48$)/33.20($\pm 0.32$) &
      66.20($\pm 0.38$)/90.70($\pm 0.30$) &
      80.70($\pm 0.42$)/89.60($\pm 0.34$) &
      N/A \\
    RotPred\cite{Hendrycks2019SSL} &
      81.12($\pm 0.33$)/92.68($\pm 0.10$) &
      60.91($\pm 0.27$)/26.49($\pm 0.13$) &
      65.07($\pm 0.30$)/92.93($\pm 0.09$) &
      84.71($\pm 0.17$)/90.52($\pm 0.11$) &
      86.81($\pm 0.18$) \\
    ARPL\cite{Chen2022ARPL} &
      82.10($\pm 0.18$)/89.05($\pm 0.21$) &
      63.80($\pm 0.34$)/39.70($\pm 0.36$) &
      67.70($\pm 0.26$)/88.10($\pm 0.30$) &
      84.05($\pm 0.28$)/87.00($\pm 0.32$) &
      84.10($\pm 0.34$) \\
    LogitNorm\cite{Wei2022LogitNorm} &
      82.42($\pm 0.22$)/93.21($\pm 0.27$) &
      57.06($\pm 0.34$)/\textbf{26.15}($\pm 0.28$) &
      67.73($\pm 0.26$)/93.44($\pm 0.22$) &
      86.22($\pm 0.30$)/92.46($\pm 0.26$) &
      86.48($\pm 0.18$) \\
    \textbf{BootOOD (Ours)} &
      \textbf{84.16}($\pm 0.28$)/89.32($\pm 0.62$) &
      \textbf{53.16}($\pm 0.22$)/33.13($\pm 0.49$) &
      \textbf{68.17}($\pm 0.30$)/89.78($\pm 0.28$) &
      \textbf{86.78}($\pm 0.65$)/87.59($\pm 0.14$) &
      \textbf{86.60}($\pm 0.15$) \\
    \midrule

    \multicolumn{6}{l}{\textbf{\textit{Training Methods w/ Outlier Data}}} \\
    UDG\cite{Yang2021SCOOD} &
      71.63($\pm 1.52$)/79.34($\pm 2.42$) &
      66.42($\pm 0.95$)/58.83($\pm 1.25$) &
      76.19($\pm 0.72$)/82.01($\pm 1.05$) &
      66.73($\pm 0.80$)/79.58($\pm 1.12$) &
      67.94($\pm 1.39$) \\
    MixOE\cite{Zhang2023MixOE} &
      80.05($\pm 0.07$)/86.83($\pm 0.42$) &
      59.61($\pm 0.21$)/47.38($\pm 0.29$) &
      \textbf{81.68}($\pm 0.18$)/88.20($\pm 0.23$) &
      72.93($\pm 0.20$)/85.94($\pm 0.27$) &
      85.01($\pm 0.10$) \\
    MCD\cite{Yu2019MCD} &
      81.42($\pm 0.16$)/86.70($\pm 0.22$) &
      61.03($\pm 0.27$)/48.12($\pm 0.30$) &
      79.85($\pm 0.22$)/86.03($\pm 0.24$) &
      70.58($\pm 0.25$)/83.84($\pm 0.27$) &
      85.12($\pm 0.26$) \\
    OE\cite{Hendrycks2019OE} &
      \textbf{85.02}($\pm 0.24$)/88.96($\pm 0.31$) &
      \textbf{52.11}($\pm 0.28$)/35.01($\pm 0.27$) &
      70.51($\pm 0.22$)/\textbf{90.34}($\pm 0.24$) &
      86.85($\pm 0.26$)/84.95($\pm 0.25$) &
      85.90($\pm 0.18$) \\
    \textbf{BootOOD (Ours)} &
       84.16($\pm 0.28$)/\textbf{89.32}($\pm 0.62$) &
       53.16($\pm 0.22$)/\textbf{33.13}($\pm 0.49$) &
       68.17($\pm 0.30$)/89.78($\pm 0.28$) &
       \textbf{86.78}($\pm 0.65$)/\textbf{87.59}($\pm 0.14$) &
       \textbf{86.60}($\pm 0.15$) \\

    \bottomrule
  \end{tabular}
  }
\end{table}

\end{document}